\documentclass[10pt,journal,cspaper,compsoc]{IEEEtran}
\usepackage{times}
\usepackage{epsfig}
\usepackage{graphicx}
\usepackage{amssymb}
\usepackage{multirow}
\usepackage{hhline}
\usepackage{booktabs}
\usepackage[fleqn]{amsmath}

\usepackage{caption}
\usepackage[ruled, vlined]{algorithm2e}

\usepackage{float}
\usepackage{bm}
\usepackage{changepage}
\usepackage{subfigure}
\usepackage{multirow}
\usepackage{array}
\usepackage{todonotes}
\usepackage{changepage}
\usepackage{url}
\usepackage{cite}

\newcommand{\keypoint}[1]{\vspace{0.2cm}\noindent\textbf{#1}\quad}
\newcolumntype{M}[1]{>{\centering\arraybackslash}m{#1}}

\ifCLASSOPTIONcompsoc
\else
\fi

\ifCLASSINFOpdf
\else
\fi

\begin{document}

\title{Zero and Few Shot Learning with Semantic Feature Synthesis and Competitive Learning}


\author{Zhiwu~Lu,~Jiechao~Guan,~Aoxue~Li,~Tao~Xiang,~An~Zhao,~and~Ji-Rong~Wen% <-this % stops a space
\thanks{Z. Lu, J. Guan, A. Zhao, and J.-R. Wen are with the Beijing Key Laboratory of Big Data Management and Analysis Methods, School of Information, Renmin University of China, Beijing 100872, China. E-mail: zhiwu.lu@gmail.com.}
\thanks{A. Li is with the School of Electronics Engineering and Computer Science, Peking University, Beijing 100871, China.}
\thanks{T. Xiang is with the School of Electronic Engineering and Computer Science, Queen Mary University of London,  London E1 4NS, and Samsung AI Centre, Cambridge,  United Kingdom. E-mail: t.xiang@qmul.ac.uk.}
}

\IEEEcompsoctitleabstractindextext{
\begin{abstract}
  Zero-shot learning (ZSL) is made possible by learning a projection function between a feature space and a semantic space (e.g.,~an attribute space). Key to ZSL is thus to learn a projection that is robust against the often large domain gap between the seen and unseen class domains. In this work, this is achieved by unseen class data synthesis and robust projection function learning. Specifically, a novel semantic data synthesis strategy is proposed, by which  semantic class prototypes (e.g., attribute vectors) are used to simply perturb seen class data for generating unseen class ones. As in any data synthesis/hallucination approach, there are ambiguities and uncertainties on how well the  synthesised data can capture the targeted unseen class data distribution. To cope with this, the second contribution of this work is a novel projection learning model termed competitive bidirectional projection learning (BPL) designed to best utilise the ambiguous synthesised data. Specifically, we assume that each synthesised data point can belong to any unseen class; and the most likely two class candidates are exploited to learn a robust projection function in a competitive fashion. As a third contribution, we show that the proposed ZSL model can be easily extended to few-shot learning (FSL) by again exploiting semantic (class prototype guided) feature synthesis and competitive BPL. Extensive experiments show that our model achieves the state-of-the-art results on both problems.
\end{abstract}


}

\maketitle

% \fi
%\IEEEpeerreviewmaketitle

%%%%%%%%% BODY TEXT
\section{Introduction}

Recently, the object recognition research has been focused on large-scale recognition problems such as the ImageNet ILSVRC challenge \cite{Russakovsky2015ImageNet}. The latest deep neural network (DNN) based models \cite{simonyan2014arxiv,szegedy2015cvpr,he2016cvpr} have achieved super-human performance on the ILSVRC 1K recognition task. A question thus naturally arises: Are we close to solving the large-scale object recognition problem? The answer clearly depends on how large the scale is: There are approximately 8.7 million animal species alone; in this context, the ILSVRC 1K recognition task is nowhere near large-scale. Importantly, existing supervised learning based methods are intrinsically limited in scalability. Specifically, they typically require hundreds of image samples to be collected from each object class. However, many object classes are rare; it is thus impossible to collect sufficient training samples for them, even with the help from social media platforms -- 296 classes in ImageNet have one image each \cite{Russakovsky2015ImageNet}. Therefore, it is  highly desirable to develop object recognition models that require only few  (say five)  or better still, zero training samples/shots  per object class.

To this end, zero-shot learning (ZSL) has become topic \cite{scheirer2013pami,Lampert2014pami,Shigeto2015,Changpinyo2016CVPR,zhang2016eccv,chao2016empirical,akata2016pami,Xian2017CVPR}. ZSL is inspired by the ability of humans in recognising unseen objects by exploiting the knowledge distilled from seen classes. For example, if a child has seen a horse before and learned from a textbook that a zebra looks very similar to a horse but has black and white stripes, s/he would then have no problem in recognising a zebra when seeing one. Similarly, to learn a ZSL model, a set of seen classes with labelled training samples are needed. In addition,  semantic descriptions of both seen and unseen classes are required so that  knowledge can be transferred from seen classes to unseen ones.

Existing ZSL models assume that each class name is embedded in a semantic space, such as attribute space \cite{kankuekul2012online,Lampert2014pami} or word vector space \cite{Frome2013nips,socher2013nips}. Given a set of seen class samples, the visual features are first extracted, typically using a DNN model pretrained on ImageNet. With the visual feature representation of the image samples and the semantic representation of the class names (termed as class prototypes), the next task is to learn a joint embedding space using the seen class training data. In such a space, both feature and semantic representations are projected so that they can be directly compared. Once the projection functions are learned, they are applied to the unseen test samples and unseen class names, and the final recognition is conducted by simple search of the nearest neighbour class prototype for each test sample.

One of the biggest challenges in ZSL is the domain gap between the seen and unseen classes. As mentioned above, the projection functions learned from the seen classes with labelled data are applied to the unseen class data in ZSL. However, the unseen classes are often visually very different from the seen ones due to the domain gap, even when they are described using, for instance, a same set of attributes (e.g., both horses and cats have tails, thus sharing the attribute `has tail'; nevertheless the visual appearance of tails can be drastically different for the two classes). Consequently, the same projection function may not be able to project an unseen class sample to be close to its corresponding class name in the joint embedding space for correct recognition. To tackle the projection domain shift \cite{fu2015transductive,Kodirov2015ICCV,rohrbach2013transfer} caused by the domain gap, a number of ZSL models resort to transductive learning \cite{guo2016aaai,ye2017cvpr,li2017tgrs,wang2017ijcv,yu2017transductive} in order to narrow the domain gap using the unlabelled unseen class samples. However, the assumption that a large number of unseen class samples are somehow collected for model training is contradictory to the problem setting of ZSL.

\begin{figure}[t]
\vspace{0.1in}
\centering
\includegraphics[width=0.90\columnwidth]{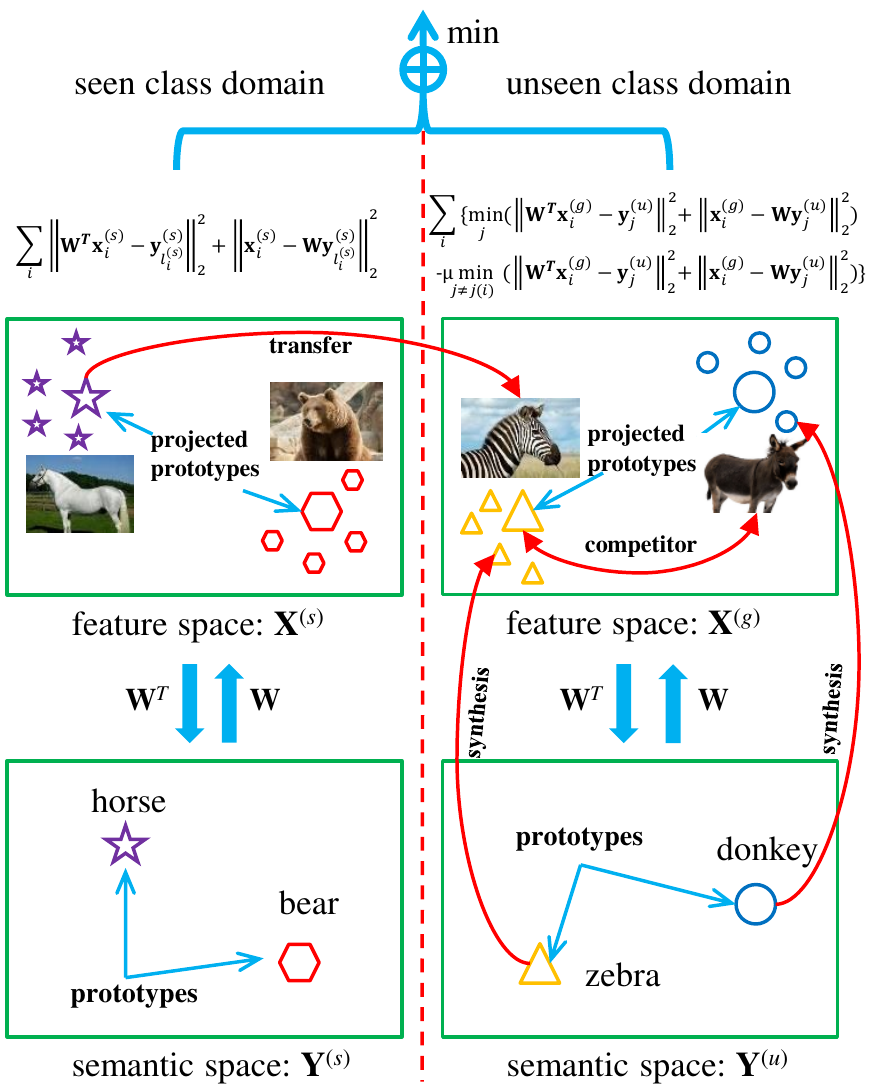}
\vspace{-0.00in}
\caption{Schematic of the proposed model.}\label{fig:pipeline}
\vspace{-0.02in}
\end{figure}

Without having access to real training samples of unseen classes, one approach to overcome the domain gap is to synthesise data from the unseen classes. Indeed, this approach has been adopted by a number of very recent ZSL models that produced state-of-the-art performance \cite{bucher2017iccv,Mishra2017,Long2017FromZL,Xian2018CVPR,Zhu2018CVPR}. Most of them utilise a generative model, often known as a generator or decoder. Once trained, a sampled random vector together with the prototype (e.g., attribute vector) of an unseen class are fed into the generator to synthesise the convolutional neural network (CNN) features of a data sample of that class. With the synthesised unseen class features, a projection model or classifier can now be directly learned with both seen and unseen classes. This approach seems to be able to remove the domain gap completely, provided that the synthesised samples are representative of the unseen class. However, it is worth noting that to train the generator, only seen classes are used. The generator itself thus suffers from the same domain gap problem and there is no guarantee that the synthesised features follow the true distribution of the unseen class specified by the prototype. The domain gap problem is thus not solved but merely embodied in a different model/form.

In this paper, we propose a novel feature synthesis method and a robust projection learning model to deal with the inevitable imperfection in the synthesised data. As the first contribution, a simple yet effective feature synthesis method is developed. Concretely, our method is based on perturbing the seen class features, guided by the semantic prototypes projected in the feature space. In this way, projection learning and feature synthesis are closely integrated. Importantly, by perturbing seen class features towards the direction of a projected unseen class prototype, representing the class centre, the intra class variations caused by factors such as pose and lighting, are preserved and transferred directly to the unseen class. This is in stark contrast to most existing works \cite{bucher2017iccv,Mishra2017,Xian2018CVPR,Zhu2018CVPR} that employ various generative models such as generative adversarial network (GAN) or variational autoencoder (VAE) that aim to capture the (seen) class distribution and implicitly transfer that to unseen classes. We believe that our much simpler and direct approach, termed as semantic feature synthesis by perturbation (SFSP), is more effective in transferring intra-class variations across domains.

No matter how effective the synthesiser is, armed with only a single prototype for an unseen class, it would inevitably be ambiguous and uncertain regarding which target unseen class the synthesised data should be assigned to. An example is shown in Fig.~\ref{fig:pipeline}. With our SFSP approach, we aim to synthesise some (unseen) zebra features using (seen) horse samples as raw material. However, the synthesised features may end up being more representative of another  unseen class, namely donkey, which is both visually and semantically similar to zebra. This is despite the fact that the feature perturbation was actually guided by the projected prototype of zebra.

To cope with the imperfect unseen class synthesis, the second contribution of the work is  a novel  bidirectional projection learning (BPL) model for learning the projection between semantic and feature spaces. The learned projection is robust against the ambiguity of the synthesised features thanks to a competitive learning formulation. As shown in Fig.~\ref{fig:pipeline}, using our competitive BPL model, both the projection from semantic to feature space and that along the opposite direction are learned (hence bidirectional). Such an autoencoder style formulation improves the model generalisation ability, as demonstrated previously in various problems  \cite{lu2013speech,ap2014autoencoder,Chen2018CVPR}. More critically, our BPL model is learned  using generalised competitive learning \cite{lu2009generalized,silva2012stochastic}. That is, unlike existing works, we do not forcefully assign a class label to the synthesised data point. Rather, we let the learned projection/classifier decide which two most likely unseen classes out of all candidates the data sample belongs to. Then in an iterative manner, we update the projection so that the updated BPL model becomes more confident/less ambiguous. In the example mentioned above, with the current projection model,  although we wanted to synthesise a zebra by recycling a horse, it turns out to be more like a donkey. This unexpected outcome is happily accepted and the synthesised data is treated as a donkey and used to update the projection so that it would not be confused to a zebra in the next iteration. Such an iterative learning strategy results in a hybrid gradient descent/ascent formulation: The most likely class label is used for gradient descent, whilst the difference between the most and second-most likely class labels is forced to be large by gradient ascent. Solving such optimisation problem is non-trivial, and an efficient iterative algorithm is formulated as the solver, followed by rigorous theoretic algorithm analysis.

As the third contribution, we show that the proposed SFSP+BPL based ZSL framework can be easily extended to solve the closely related few-shot learning (FSL) problem \cite{snell2017prototypical,sung2017learning,qiao2017few}. Early solutions are dominated by meta-learning \cite{finn2017model,santoro2016meta,sung2017meta,andrychowicz2016learning}, but as in ZSL, feature synthesis based FSL \cite{hariharan2017low,Douze_2018_CVPR,Wang_2018_CVPR} starts to show superior performance. With a few (e.g., five) shots from a target class, the same SFSP strategy is adopted, with the only difference that the few shots rather than the seen (source) class samples are used as the raw material for synthesis. We thus provide a unified approach to both zero and few shot learning. Extensive experiments are carried out using benchmarks for both ZSL and FSL. The results show that our approach outperforms the state-of-the-art alternatives, often by significant margins.

\vspace{-0.1cm}
\section{Related Work}
\label{sect:related_work}

\vspace{-0.1cm}
\keypoint{Projection Learning for ZSL}
Existing ZSL models fall into three groups, depending on how the project functions that align the visual feature and semantic spaces are learned: (1) The first group of models learn a projection function from a visual feature space to a semantic space (i.e. in a forward projection direction) by employing conventional regression/ranking models \cite{Lampert2014pami,Akata2015CVPR} or deep neural network regression/ranking models \cite{socher2013nips,Frome2013nips,lei2015predicting}. (2) The second group of models choose the reverse projection direction \cite{Shigeto2015,Kodirov2015ICCV,shojaee2016semi,Zhang2017cvpr}, i.e., from the semantic space to the feature space, to alleviate the hubness problem suffered by nearest neighbour search in a high dimensional space \cite{radovanovic2010hubs}. (3) The third group of models learn an intermediate space as the embedding space, where both the feature space and the semantic space are projected to \cite{lu2015unsupervised,zhang2016cvpr,Changpinyo2016CVPR}. Our BPL model integrates the forward and reverse projections. This autoencoder style bidirectional projection learning strategy has been adopted by a number of recent ZSL models \cite{Kodirov2017CVPR,Chen2018CVPR} to improve the generalisability of the learned projection to the unseen class domain. However, none of them additionally exploits unseen class feature synthesis to tackle the projection domain gap problem explicitly.

\keypoint{Data Synthesis Based ZSL}
To cope with the extreme data imbalance problem in ZSL (zero training samples for unseen classes and plenty for seen classes), a number of recent studies exploit the idea of synthesising unseen class samples \cite{bucher2017iccv,Mishra2017,Long2017FromZL,Xian2018CVPR,Zhu2018CVPR}. In \cite{Long2017FromZL}, a discriminative model is learned to map a class prototype into a data feature vector. Such a mapping is deterministic, i.e., one unseen class prototype can only be used to synthesise one visual feature, hardly helping solve the imbalance problem at all. So the model uses unseen class instance (per image) attribute vectors instead of per class attribute vector as input. This clearly violates the ZSL setting: how can one collect unseen class sample attributes when the images are not available? All other studies employ a generative model which samples a random vector and combines that with the unseen class prototype to form the input to a generator. In this way, an arbitrary number of unseen class samples can be synthesised. They differ mainly in the generative models used: ranging from GAN \cite{Xian2018CVPR}, VAE \cite{Mishra2017}, to Generative Moment Matching Network (GMMN) \cite{bucher2017iccv}. Our data synthesis method differs significantly from the existing ones in both how data are synthesised and once synthesised, how they are used for ZSL model training. First, our unseen class data are synthesised by directly perturbing the seen class samples towards the direction of the projected unseen class prototypes. Data perturbation has been widely used for data argumentation in supervised learning for seen classes, but never been used for unseen class synthesis before. Importantly by avoiding the challenging generative model training process and directly transferring the intra-class variation from seen to unseen classes, our method becomes more effective in narrowing down the domain gap. Second, instead of assigning a fixed class label to a synthesised sample, we allow it to be assigned to any classes that are deemed plausible based on the current projection model. By accepting that the synthesised data class membership is ambiguous and developing a learning strategy (competitive learning) to cope with it, our model is more robust against any imperfection in data synthesis.

%\vspace{-0.1cm}
\keypoint{Few-Shot Learning}
As a related problem to ZSL, few-shot learning (FSL) \cite{snell2017prototypical,sung2017learning,qiao2017few} assumes that a handful (typically 1-5) labelled examples exist for target/novel classes. Such a data sparsity issue challenges the standard fine-tuning strategy used in deep learning. Data augmentation can alleviate the issue, but does not solve it. Recent FSL approaches thus choose to transform the deep network training process to meta-learning where the transferrable knowledge is learned in the form of good initial conditions, embeddings, or optimisation strategies \cite{santoro2016meta,finn2017model,sung2017meta,andrychowicz2016learning}. However,  meta-learning based models typically cannot scale to large number of classes/training samples. This motivated the more recent approaches based on classifier parameter prediction with activations \cite{qiao2017few} and data synthesis based FSL \cite{hariharan2017low,Douze_2018_CVPR,Wang_2018_CVPR}. In this work, our ZSL model is seamlessly extended to FSL. Since it is data synthesis based, it is related to \cite{hariharan2017low,Douze_2018_CVPR,Wang_2018_CVPR}, but is again distinctive in the semantic feature synthesis by pertubation (SFSP) strategy and its ability to cope with the ambiguity in the synthesised data.  We show that our model significantly outperforms the state-of-the-art FSL alternatives \cite{qiao2017few,hariharan2017low}, particularly on large-scale FSL tasks (see Fig.~\ref{fig:metafsl1}).

%\vspace{-0.1cm}
\keypoint{Gradient-Based Optimisation}
A hybrid gradient descent/ascent algorithm is developed to train our competitive BPL model, inspired by generalised competitive learning \cite{lu2009generalized,silva2012stochastic}. This formulation yields a mix of min-min and max-min optimisation problems. Solving max-min (or min-min) problems is non-trivial, and many complicated optimisation methods have been developed such as entropy-based aggregate method \cite{xingsi1992entropy} and projected Lagrangian \cite{murray1980projected}. In this paper, we propose a gradient-based optimisation algorithm which differs from existing optimisation methods in that: the gradient descent and ascent strategies are combined in a unified framework, resulting in an extremely efficient solver with rigorous theoretic algorithm analysis provided.

\keypoint{Earlier Version of the Work} An earlier and preliminary version of this work is published in \cite{Zhao2018nips}. A number of significant modifications have been made in this version: (1) Data synthesis based on seen class feature perturbation is introduced to turn the ZSL framework from a transductive one into an inductive one. (2) Competitive learning is incorporated into the BPL model to improve its robustness and generalisability. (3) An extension to FSL is formulated.

\vspace{-0.1cm}
\section{Zero-Shot Learning}
\label{sect:method}

\vspace{-0.0cm}
\subsection{Problem Definition}

Let $\mathcal{S}=\{s_1,...,s_p\}$ denote a set of seen classes and $\mathcal{U}=\{u_1,...,u_q\}$ denote a set of unseen classes, where $p$ and $q$ are the total numbers of seen  and unseen classes, respectively. These two sets of classes are disjoint, i.e. $\mathcal{S}\cap \mathcal{U}=\phi$. Similarly, $\mathbf{Y}_s = [\mathbf{y}_1^{(s)},..., \mathbf{y}_p^{(s)}] \in \mathbb {R}^{k\times p}$ and $\mathbf{Y}_u = [\mathbf{y}_1^{(u)},..., \mathbf{y}_q^{(u)}] \in \mathbb{R}^{k\times q}$ denote the corresponding seen and unseen class semantic representations/prototypes (e.g., $k$-dimensional attribute vector). We are given a set of labelled training samples $\mathcal{D}_s=\{(\mathbf{x}_i^{(s)}, l_i^{(s)}, \mathbf{y}_{l_i^{(s)}}^{(s)}): i=1,...,N_s\}$, where $\mathbf{x}_i^{(s)} \in \mathbb{R}^{d\times1}$ is the $d$-dimensional visual feature vector of the $i$-th sample in the training set, $l_i^{(s)} \in\{1,...,p\}$ is the label of $\mathbf{x}_i^{(s)}$ according to $\mathcal{S}$, $\mathbf{y}_{l_i^{(s)}}^{(s)}$ is the semantic representation of $\mathbf{x}_i^{(s)}$, and $N_s$ denotes the total number of labelled samples. Let $\mathcal{D}_u= \{(\mathbf{x}_i^{(u)}, l_i^{(u)}, \mathbf{y}_{l_i^{(u)}}^{(u)}): i=1,...,N_u\}$ be a set of unlabelled test samples, where $\mathbf{x}_i^{(u)} \in \mathbb{R}^{d\times1}$ is the $d$-dimensional visual feature vector of the $i$-th sample in the test set, $l_i^{(u)} \in\{1,...,q\}$ is the unknown label of $\mathbf{x}_i^{(u)}$ according to $\mathcal{U}$, $\mathbf{y}_{l_i^{(u)}}^{(u)}$ is the unknown semantic representation of $\mathbf{x}_i^{(u)}$, and $N_u$ denotes the total number of unlabelled samples. The goal of zero-shot learning is to predict the labels of test samples by learning a classifier $f:\mathcal{X}_u\rightarrow \mathcal{U}$, where $\mathcal{X}_u = \{\mathbf{x}_i^{(u)}: i=1,...,N_u\}$. In a generalised setting, the test samples can come from both seen and unseen classes, so the classifier becomes $f:\mathcal{X}\rightarrow \mathcal{S} \cup \mathcal{U}$, where $\mathcal{X}$ denotes all test samples.

\vspace{-0.1cm}
\subsection{Semantic Feature Synthesis by Perturbation}
\label{sec:ZSL feature synthesis}

We first describe how unseen class data are synthesised using our Semantic Feature Synthesis by Perturbation (SFSP) strategy.  Let $\mathbf{W} \in \mathbb{R}^{d\times k}$ be an initial projection matrix learned by any linear projection learning model (e.g. \cite{Kodirov2017CVPR}) with all seen class samples. With $\mathbf{W}$, a class prototype can be projected into the feature space to act as a class centre.  Using SFSP, for a given unseen class, we choose some semantically related (i.e., close in the semantic space) seen classes and perturb their sample features towards the unseen class centre to synthesise unseen class samples. Formally, given  $q$ unseen class prototypes $\mathbf{Y}_u = [\mathbf{y}_1^{(u)},..., \mathbf{y}_q^{(u)}]$, a set of unseen class features are synthesised as:
\begin{equation}
\begin{small}
{\mathbf{x}}_i^{(g)} = {\mathbf{x}}_i^{(s)} + \rho \frac{\mathbf{W}(\mathbf{y}_j^{(u)}- \mathbf{y}_{j'}^{(s)} )}{\|\mathbf{W}\|^2_F}~(1\leq j\leq q), \label{eq:generate}
\end{small}
\end{equation}
where $\mathbf{y}_{j'}^{(s)}$ falls in the $k_g$-nearest neighbours of $\mathbf{y}_j^{(u)}$ among  all seen classes in the semantic space (e.g., attribute space), ${\mathbf{x}}_i^{(s)}$ is randomly selected from the $j'$-th seen class (its prototype is $\mathbf{y}_{j'}^{(s)}$), and $\rho$ is a weight parameter in the range $(0,1)$.  In this work, we empirically set the neighbourhood size $k_g$ to 3 and the number of randomly selected samples from each seen class in the neighbourhood to 15. All synthesised unseen class samples are collected into a set $\mathcal{X}_g = \{\mathbf{x}_i^{(g)}: i=1,...,N_g\}$, where $N_g$ is the total number of generated samples and $N_g =15\times q$ in this work.

\begin{figure}[t]
\vspace{0.08in}
\centering
\includegraphics[width=0.98\columnwidth]{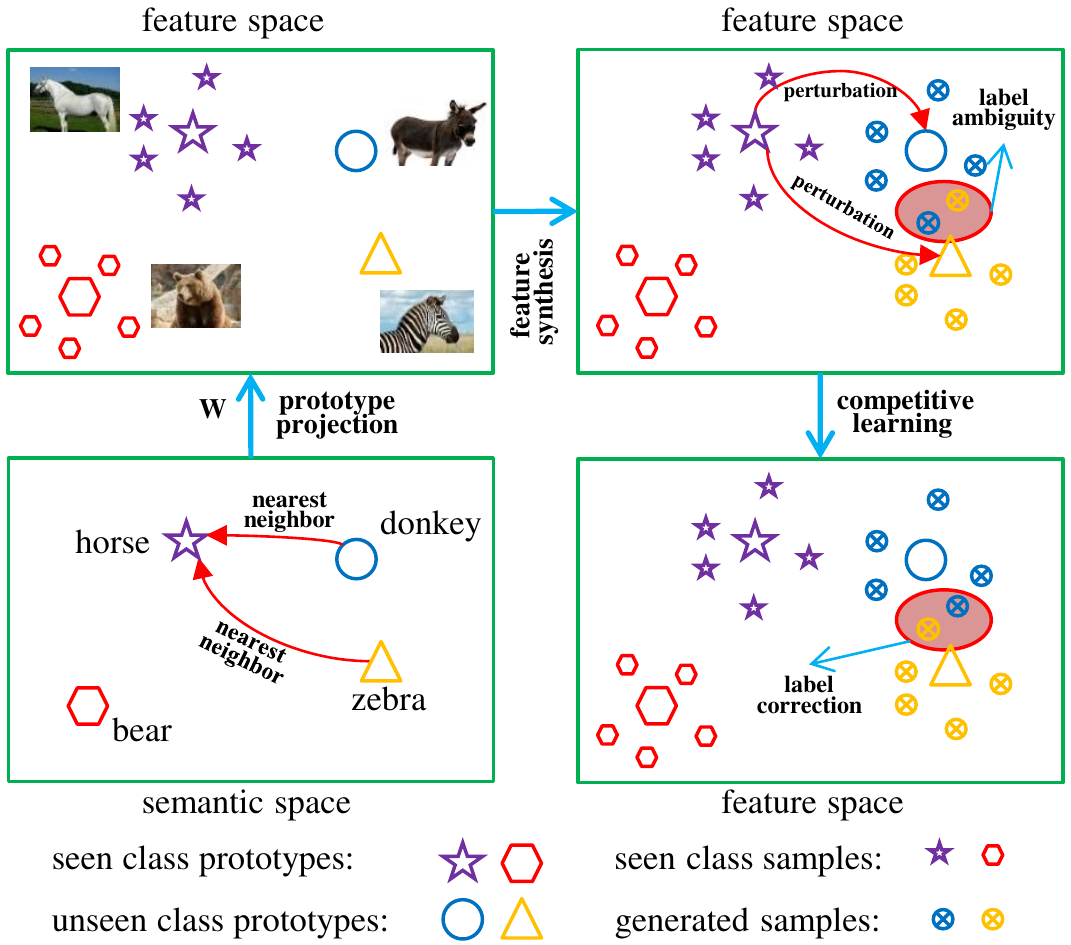}
\vspace{0.02in}
\caption{Illustration of our SFSP strategy used for ZSL.}\label{fig:genzsl}
\vspace{-0.00in}
\end{figure}

Note that our SFSP strategy is formulated based on the assumption that semantically similar classes (their prototypes being close in the semantic space) have similar local graph structures in the feature space. This assumption is reasonable since it is well known that the final layer of a CNN model is often semantically meaningful and abstract. This is why according to Eq.~(\ref{eq:generate}), the class centre offset ($\mathbf{y}_j^{(u)}- \mathbf{y}_{j'}^{(s)}$) is used as the direction of perturbation in the feature space. Moreover, different from existing generative models \cite{bucher2017iccv,Mishra2017,Chen2018CVPR,Xian2018CVPR}, even when $\mathbf{y}_j^{(u)}$ is used, the synthesised feature vector is not assumed to belong to the $j$-th unseen class. The process of determining which class label should be assigned, termed `label correction' (illustrated in Fig.~\ref{fig:genzsl}), is coupled with the learning of the projection function, to be detailed next.

\vspace{-0.1cm}
\subsection{Competitive Bidirectional Projection Learning}

We aim to learn a projection function (denoted by $\mathbf{W}$) that can project a class prototype in a semantic space to a (CNN) feature space. To this end, we employ bidirectional projection learning (BPL) in a competitive learning formulation. Specifically, to perform BPL with the labelled seen class samples $\mathcal{D}_s$ and generated unseen class samples $\mathcal{X}_g$, our model solves the following optimisation problem:
\setlength{\arraycolsep}{0.1em}
\begin{small}
\begin{eqnarray}
&&\hspace{-0.06in}\min_{\mathbf{W}}\hspace{-0.02in}\sum_{i=1}^{N_s} (\|\mathbf{W}^T \mathbf{x}_i^{(s)}\hspace{-0.02in} - \mathbf{y}_{l_i^{(s)}}^{(s)}\|^2_2 + \|\mathbf{x}_i^{(s)}\hspace{-0.02in} - \mathbf{W}\mathbf{y}_{l_i^{(s)}}^{(s)}\|^2_2) + 2\nu \|\mathbf{W}\|^2_F \nonumber \\
&&\hspace{0.15in}+ \gamma\sum_{i=1}^{N_g} \{\min_j(\|\mathbf{W}^T{\mathbf{x}}_i^{(g)} \hspace{-0.02in}- \mathbf{y}_j^{(u)}\|^2_2 + \|{\mathbf{x}}_i^{(g)} \hspace{-0.02in}- \mathbf{W}\mathbf{y}_j^{(u)}\|^2_2) \nonumber\\
&&\hspace{0.25in} -\mu \min_{j\neq j(i)}(\|\mathbf{W}^T{\mathbf{x}}_i^{(g)} \hspace{-0.02in}- \mathbf{y}_j^{(u)}\|^2_2 + \|{\mathbf{x}}_i^{(g)} \hspace{-0.02in}- \mathbf{W}\mathbf{y}_j^{(u)}\|^2_2)\},
\label{eq:erplobj}
\end{eqnarray}
\end{small}
\hspace{-5pt}where $j(i)=\arg\min_j(\|\mathbf{W}^T{\mathbf{x}}_i^{(g)} \hspace{-0.02in}- \mathbf{y}_j^{(u)}\|^2_2 + \|{\mathbf{x}}_i^{(g)} \hspace{-0.02in}- \mathbf{W}\mathbf{y}_j^{(u)}\|^2_2)$, $\gamma$ is a weight parameter that controls the importance of the losses on the seen and unseen class samples, and $\mu$ is a weight parameter that corresponds to the strength of generalised competitive learning on the unseen class samples. Note that althought two projection directions are involved, the two projection matrices are transpose of each other, similar to those in an autoencoder \cite{baldi2012autoencoders}.

There are three terms in Eq.~(\ref{eq:erplobj}). The first term (1st row of Eq.~(\ref{eq:erplobj})) combines the forward and reverse projection errors on the seen class training samples. The second term (also 1st row of Eq.~(\ref{eq:erplobj})) is a regularisation term on $\mathbf{W}$ with a weight of $\nu$. The third term (2nd and 3rd rows of Eq.~(\ref{eq:erplobj})) are the training objective computed on the synthesised unseen class samples $\mathcal{X}_g$ and is the place where competitive learning takes place. Specifically, it is formulated by subtracting the minimum function with respect to the second-best label for ${\mathbf{x}}_i^{(g)}$ (3rd row) from the minimum function with respect to the best label for ${\mathbf{x}}_i^{(g)}$ (2nd row). Essentially, this term dictates that each synthesised sample will be close to the most likely unseen class centre, whilst being farther away to the second best, consequently the rest of the candidate unseen class centres. This is essentially an unsupervised learning objective. It is appropriate for our case because, due to the ambiguity of the data synthesis process, we are not certain which unseen class each synthesised data point should belong to. Instead, we enforce that it should belong to one of the unseen classes and therefore should be pushed away from the rest (see Fig.~\ref{fig:genzsl}). It is worth noting that, although our model only has a simple linear formulation, it is \emph{clearly shown to outperform} existing nonlinear related autoencoder-based models \cite{Wang2018AAAI,Mishra2017} (see Table~\ref{tab:pzsl}).

\vspace{-0.1cm}
\subsection{Optimisation}
\label{sec:alg}

The learning problem defined in Eq.~(\ref{eq:erplobj}) consists of a mix of min-min and max-min subproblems; solving it is thus non-trivial. In the following, by employing a hybrid gradient descent/ascent strategy, a gradient-based solver is developed, resulting in an efficient iterative algorithm.

Given the projection matrix $\mathbf{W}^{(t)}$ at iteration $t$, we define the loss function $\mathbf{f}_i^{(t)} = [f_{i1}^{(t)},...,f_{iq}^{(t)}]^T$ for the synthesised unseen class sample ${\mathbf{x}}_i^{(g)}~(i=1,...,N_g)$, where $f_{ij}^{(t)} = \|{\mathbf{W}^{(t)}}^T{\mathbf{x}}_i^{(g)} - \mathbf{y}_j^{(u)}\|^2_2 + \|{\mathbf{x}}_i^{(g)} - \mathbf{W}^{(t)} \mathbf{y}_j^{(u)}\|^2_2 ~(j=1,...,q)$. For the minimum function $\min_j \mathbf{f}_i^{(t)}= \min_j [f_{ij}^{(t)}]_{q\times1}$, its gradient can be written as $\eta_i^{(t)} =[\eta_{i1}^{(t)}, ..., \eta_{iq}^{(t)}]^T$ with respect to $\mathbf{f}_i^{(t)}$ as in our earlier and preliminary work \cite{Zhao2018nips}:
\begin{eqnarray}
\eta_{ij}^{(t)} =
\left\{
\begin{array}{cl}
1/n_i^{(t)} &,~~\mathrm{if}~f_{ij}^{(t)}=\min_j\mathbf{f}_i^{(t)} \\
0       &,~~\textrm{otherwise}
\end{array},
\right. \label{eq:subgrad0}
\end{eqnarray}
where $n_i^{(t)}$ is the number of $f_{ij}^{(t)}~(j=1,...,q)$ equalling to $\min_j \mathbf{f}_i^{(t)}$. Similarly, for the minimum function $\min_{j\neq j(i)}\mathbf{f}_i^{(t)}=\min_{j\neq j(i)}[f_{ij}^{(t)}]_{q\times1}$, we define its gradient $\xi_i^{(t)} =[\xi_{i1}^{(t)}, ..., \xi_{iq}^{(t)}]^T$ with respect to $\mathbf{f}_i^{(t)}$:
\begin{eqnarray}
\xi_{ij}^{(t)} =
\left\{
\begin{array}{cl}
1/m_i^{(t)} &,~~\mathrm{if}~f_{ij}^{(t)}=\min_{j\neq j(i)}\mathbf{f}_i^{(t)} \\
0       &,~~\textrm{otherwise}
\end{array},
\right. \label{eq:subgrad1}
\end{eqnarray}
where $m_i^{(t)}$ is the number of $f_{ij}^{(t)}~(j=1,...,q)$ that equals to $\min_{j\neq j(i)}\mathbf{f}_i^{(t)}$. Taking the Taylor expansion, we have:
\begin{small}
\begin{eqnarray}
& & \hspace{-0.12in} \min_j(\|{{\mathbf{W}}^{(t+1)}}^T{\mathbf{x}}_i^{(g)} - \mathbf{y}_j^{(u)}\|^2_2 + \|{\mathbf{x}}_i^{(g)} - \mathbf{W}^{(t+1)} \mathbf{y}_j^{(u)}\|^2_2) \nonumber\\
& =&\min\mathbf{f}_i^{(t+1)} \approx \min\mathbf{f}_i^{(t)}+{\eta_i^{(t)}}^T (\mathbf{f}_i^{(t+1)}-\mathbf{f}_i^{(t)})\nonumber\\
&=& (\min\mathbf{f}_i^{(t)}-{\eta_i^{(t)}}^T \mathbf{f}_i^{(t)}) +{\eta_i^{(t)}}^T \mathbf{f}_i^{(t+1)} = {\eta_i^{(t)}}^T \mathbf{f}_i^{(t+1)}.
\end{eqnarray}
\end{small}
\hspace{-4pt}Similarly, we have another approximation:
\begin{small}
\begin{eqnarray}
& & \hspace{-0.12in} \min_{j\neq j(i)}(\|{{\mathbf{W}}^{(t+1)}}^T{\mathbf{x}}_i^{(g)} - \mathbf{y}_j^{(u)}\|^2_2 + \|{\mathbf{x}}_i^{(g)} - \mathbf{W}^{(t+1)} \mathbf{y}_j^{(u)}\|^2_2) \nonumber\\
& =&\min_{j\neq j(i)}\mathbf{f}_i^{(t+1)} \approx {\xi_i^{(t)}}^T \mathbf{f}_i^{(t+1)}.
\end{eqnarray}
\end{small}
\hspace{-4pt}According to the above approximations, the objective function in Eq.~(\ref{eq:erplobj}) at iteration $t+1$ can be estimated as:
\begin{small}
\begin{eqnarray}
&&\mathcal{F}(\mathbf{W}^{(t+1)}) \nonumber\\
&=& \sum_{i=1}^{N_s} (\|{\mathbf{W}^{(t+1)}}^T \mathbf{x}_i^{(s)} -  \mathbf{y}_{l_i^{(s)}}^{(s)}\|^2_2  +\|\mathbf{x}_i^{(s)} - \mathbf{W}^{(t+1)} \mathbf{y}_{l_i^{(s)}}^{(s)}\|^2_2) \nonumber\\
&& + 2\nu \|\mathbf{W}^{(t+1)}\|^2_F + \gamma \sum_{i=1}^{N_g} (\eta_i^{(t)}-\mu \xi_i^{(t)})^T \mathbf{f}_i^{(t+1)}.
\end{eqnarray}
\end{small}

Let $\delta_i^{(t)} = [\delta_{ij}^{(t)}]_{q\times 1} = \eta_{i}^{(t)} - \mu\xi_{i}^{(t)}$. By setting $\frac{\partial \mathcal{F} (\mathbf{W}^{(t+1)})} {\partial \mathbf{W}^{(t+1)}} =0$, we obtain a linear equation:
\begin{small}
\begin{eqnarray}
\hspace{-0.25in} && \hspace{0.45in} \mathbf{A}^{(t)} \mathbf{W}^{(t+1)} + \mathbf{W}^{(t+1)} \mathbf{B}^{(t)} = \mathbf{C}^{(t)}, \label{eq:erpllinear} \\
\hspace{-0.25in} && \mathbf{A}^{(t)} = \sum_{i=1}^{N_s} \mathbf{x}_i^{(s)} {\mathbf{x}_i^{(s)}}^T + \gamma  \sum_{i=1}^{N_g} (1-\mu)\mathbf{x}_i^{(g)} {\mathbf{x}_i^{(g)}}^T + \nu I , \\
\hspace{-0.25in} && \mathbf{B}^{(t)} = \sum_{i=1}^{N_s} \mathbf{y}_{l_i^{(s)}}^{(s)} {\mathbf{y}_{l_i^{(s)}}^{(s)}}^T +\gamma  \sum_{i=1}^{N_g} \sum_{j=1}^q \delta_{ij}^{(t)}\mathbf{y}_{j}^{(u)} {\mathbf{y}_{j}^{(u)}}^T + \nu I, \\
\hspace{-0.25in}&& \mathbf{C}^{(t)} = 2\sum_{i=1}^{N_s} \mathbf{x}_{i}^{(s)} {\mathbf{y}_{l_i^{(s)}}^{(s)}}^T + 2\gamma \sum_{i=1}^{N_g} \sum_{j=1}^q \delta_{ij}^{(t)} \mathbf{x}_{i}^{(g)}{\mathbf{y}_{j}^{(u)}}^T.
\end{eqnarray}
\end{small}
\hspace{-4pt}Let $\alpha_t = \gamma/(1+\gamma) \in (0, 1)$ and $\beta = \nu/(1+\gamma)$. In this work, we empirically set $\alpha_t=0.99^t\alpha~(\alpha \in (0, 1))$ and $\beta=0.01$. We thus have:
\begin{small}
\begin{eqnarray}
\hspace{-0.2in} \widehat{\mathbf{A}}^{(t)} &=& (1\hspace{-0.05in} -\hspace{-0.03in}\alpha_t) \hspace{-0.03in} \sum_{i=1}^{N_s} \mathbf{x}_i^{(s)} {\mathbf{x}_i^{(s)}}^T \hspace{-0.08in} + \hspace{-0.03in} \alpha_t \hspace{-0.03in} \sum_{i=1}^{N_g} (1\hspace{-0.03in} - \hspace{-0.03in} \mu)\mathbf{x}_i^{(g)} {\mathbf{x}_i^{(g)}}^T \hspace{-0.08in}+ \hspace{-0.03in} \beta I, \label{eq:erplAt}\\
\hspace{-0.2in} \widehat{\mathbf{B}}^{(t)} &=& (1\hspace{-0.05in}- \hspace{-0.03in}\alpha_t) \hspace{-0.03in} \sum_{i=1}^{N_s} \mathbf{y}_{l_i^{(s)}}^{(s)} {\mathbf{y}_{l_i^{(s)}}^{(s)}}^T \hspace{-0.08in} + \hspace{-0.03in} \alpha_t \hspace{-0.03in} \sum_{i=1}^{N_g} \sum_{j=1}^q \delta_{ij}^{(t)}\mathbf{y}_{j}^{(u)} {\mathbf{y}_{j}^{(u)}}^T \hspace{-0.08in} + \hspace{-0.03in} \beta I, \label{eq:erplBt} \\
\hspace{-0.2in} \widehat{\mathbf{C}}^{(t)} &=& 2(1\hspace{-0.05in} -\hspace{-0.03in} \alpha_t) \hspace{-0.03in} \sum_{i=1}^{N_s} \mathbf{x}_{i}^{(s)} {\mathbf{y}_{l_i^{(s)}}^{(s)}}^T \hspace{-0.08in} +\hspace{-0.03in} 2\alpha_t \hspace{-0.03in} \sum_{i=1}^{N_g} \sum_{j=1}^q \delta_{ij}^{(t)} \mathbf{x}_{i}^{(g)}{\mathbf{y}_{j}^{(u)}}^T. \label{eq:erplCt}
\end{eqnarray}
\end{small}
\hspace{-4pt}The linear equation in Eq.~(\ref{eq:erpllinear}) is then reformulated as:
\begin{small}
\begin{eqnarray}
\widehat{\mathbf{A}}^{(t)} \mathbf{W}^{(t+1)} + \mathbf{W}^{(t+1)} \widehat{\mathbf{B}}^{(t)} = \widehat{\mathbf{C}}^{(t)},  \label{eq:erpllinear1}
\end{eqnarray}
\end{small}
\hspace{-3pt}which is a Sylvester equation and it can be solved efficiently by the Bartels-Stewart algorithm \cite{BS72}.

Considering that the predicted labels of generated unseen class samples are inevitably noisy, we choose to estimate the number $n_i^{(t)}$ used in Eq.~(\ref{eq:subgrad0}) under a looser condition and redefine the gradient as follows:
\begin{equation}
\begin{small}
\eta_{ij}^{(t)} =
\left\{
\begin{array}{cl}
1/{n}_i^{(t)} &,~~\mathrm{if}~\frac{f_{ij}^{(t)}- \min\mathbf{f}_i^{(t)}}{\min\mathbf{f}_i^{(t)}}<\epsilon \\
0       &,~~\textrm{otherwise}
\end{array},
\right. \label{eq:subgrad01}
\end{small}
\end{equation}
where ${n}_i^{(t)}$ denotes the number of elements in the set $j(i)=\{j: (f_{ij}^{(t)}- \min\mathbf{f}_i^{(t)})/ {\min\mathbf{f}_i^{(t)}}<\epsilon, j=1,...,q\}$. Similarly, we redefine the gradient in Eq.~(\ref{eq:subgrad1}) as:
\begin{equation}
\begin{small}
\xi_{ij}^{(t)} =
\left\{
\begin{array}{cl}
1/{m}_i^{(t)} &,~~\mathrm{if}~\frac{f_{ij}^{(t)}- \min_{j\notin j(i)}\mathbf{f}_i^{(t)}}{\min_{j\notin  j(i)}\mathbf{f}_i^{(t)}}<\epsilon, j\notin j(i)  \\
0       &,~~\textrm{otherwise}
\end{array},
\right. \label{eq:subgrad11}
\end{small}
\end{equation}
where ${m}_i^{(t)}$ is the number of $f_{ij}^{(t)}~(j\notin j(i))$ satisfying $(f_{ij}^{(t)}- \min_{j\notin j(i)}\mathbf{f}_i^{(t)})/{\min_{j\notin  j(i)}\mathbf{f}_i^{(t)}}<\epsilon$.
In this work, we empirically set $\epsilon=0.001$ in all experiments.

\begin{algorithm}[t]
\begin{small}
  \caption{Competitive BPL}
  \label{alg:erpl}
  \SetAlgoLined
  \KwIn{~Labelled seen class samples $\mathcal{D}_s$  \\\qquad ~~~
  Synthesised unseen class samples $\mathcal{X}_g$  \\\qquad ~~~~Semantic prototypes
  $\mathbf{Y}_s,\mathbf{Y}_u$ \\~~~~\qquad Parameters $\alpha,\mu$.}
  \KwOut{$\mathbf{W}^*$}
  1. Set $t=0$\;
  2. Initialise $\mathbf{W}^{(0)}$ with our BPL model ($\alpha=0$)\;
  \While{a stopping criterion is not met}{
    3. Set $\alpha_t=0.99^t\alpha$\;
    4. Compute $\eta_{ij}^{(t)}$ and $\xi_{ij}^{(t)}$ with Eqs.~(\ref{eq:subgrad01}) and (\ref{eq:subgrad11}) \;
    5. Update $\delta_{ij}^{(t)} = \eta_{ij}^{(t)} - \mu\xi_{ij}^{(t)}$\;
    6. Compute $\widehat{\mathbf{A}}^{(t)}$, $\widehat{\mathbf{B}}^{(t)}$, and $\widehat{\mathbf{C}}^{(t)}$ with Eqs.~(\ref{eq:erplAt})--(\ref{eq:erplCt})\;
    7. Update $\mathbf{W}^{(t+1)}$ by solving Eq.~(\ref{eq:erpllinear1})\;
    8. Set $t=t+1$\;
  }
  9. return $\mathbf{W}^*=\mathbf{W}^{(t)}$.
  \end{small}
\end{algorithm}

The proposed competitive BPL algorithm is summarised in Algorithm~\ref{alg:erpl}, and a rigorous theoretic algorithm analysis can be found in the Supplementary Materials. Note that any suitable projection learning model can be used to obtain the initial projection matrix $\mathbf{W}^{(0)}$. In this paper, we choose our BPL model with $\alpha=0$ for this initialisation (i.e., BPL with seen class samples only). Once learned, given the optimal $\mathbf{W}^*$ found by our competitive BPL algorithm, we predict the label of a test sample $\mathbf{x}_i^{(u)}$ as follows:
\begin{small}
\begin{eqnarray}
\hspace{-0.15in}l_i^{(u)} \hspace{-0.03in} = \hspace{-0.03in} \arg\min_j \|{\mathbf{x}}_i^{(u)} \hspace{-0.05in} - \hspace{-0.03in} \mathbf{W^*}\mathbf{y}_j^{(u)}\|^2_2. \label{eq:abplclass}
\end{eqnarray}
\end{small}
\vspace{-0.3cm}

The time complexity analysis of our competitive BPL algorithm is given as follows. First, the computation of $[\delta_{ij}^{(t)}]_{N_g\times q}$, $\widehat{\mathbf{A}}^{(t)}$, $\widehat{\mathbf{B}}^{(t)}$, and $\widehat{\mathbf{C}}^{(t)}$ has a time complexity of $O(qN_g)$, $O(d^2(N_s+N_g))$, $O(k^2N_s+k^2N_g)$, and $O(dkN_s+dkN_g)$, respectively. Here, the sparsity of $[\delta_{ij}^{(t)}]$ is used to reduce the cost of computing $\widehat{\mathbf{B}}^{(t)}$ and $\widehat{\mathbf{C}}^{(t)}$. Second, given $\widehat{\mathbf{A}}^{(t)} \in \mathbb{R}^{d\times d}$ and $\widehat{\mathbf{B}}^{(t)} \in \mathbb{R}^{k\times k}$, the time complexity of solving Eq.~(\ref{eq:erpllinear1}) is $O(d^3+k^3)$. To sum up, one iteration of our  algorithm has a linear time complexity of $O(qN_g + (d^2+dk+k^2)(N_s+N_g)) ~(d,k,q\ll (N_s+N_g))$ with respect to the data size. We find that empirically our algorithm converges very quickly ($t\leq 5$), making it  efficient for large-scale problems.

\vspace{-0.00cm}
\section{Few-Shot Learning}

\vspace{-0.00cm}
\subsection{Problem Setting and Model Overview}
\label{sect:lsfsl}

It is straightforward to extend our ZSL model to few-shot learning (FSL), where instead of having only class prototypes and no training samples, we assume that a handful of samples do exist for the target classes. We are particularly interested in large-scale (e.g., ImageNet scale) FSL problems \cite{snell2017prototypical,sung2017learning,qiao2017few}. Here, we formulate our model with a specific large-scale FSL problem setting. The ImageNet ILSVRC2012/2010 \cite{Russakovsky2015ImageNet} (ImNet) dataset is considered, which is organised into three parts: a training set of many labelled source/base class samples, a support set of few labelled target/novel class samples, and a test set of the rest target/novel class samples. Concretely, the 1,000 classes of ILSVRC2012 are used as the base classes, and the 360 classes of ILSVRC2010 (not included in ILSVRC2012) are used as the novel classes.

\begin{figure}[t]
\vspace{0.1in}
\centering
\includegraphics[width=0.96\columnwidth]{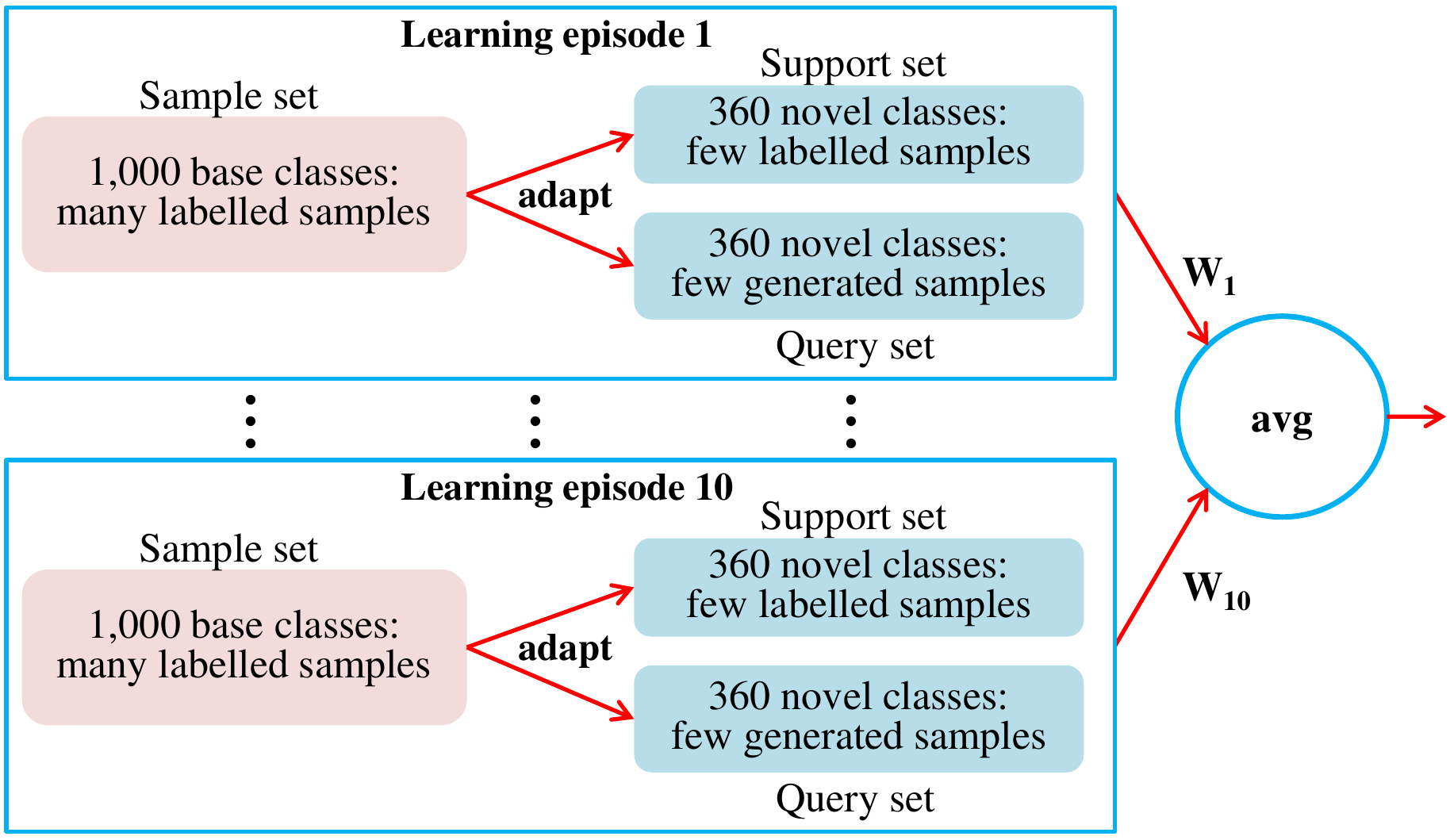}
\vspace{0.02in}
\caption{Schematic of the proposed large-scale FSL model applied to the ImNet dataset.}\label{fig:lsfsl}
\vspace{-0.00in}
\end{figure}

As illustrated in Fig.~\ref{fig:lsfsl}, to apply our competitive BPL algorithm to large-scale FSL, we create $K_e$ ($K_e=10$ in this work) learning episodes. For each episode, three sets of training samples are needed: (1) a sampled subset (i.e. sample set) of the labelled samples from the 1,000 base classes; (2) a support set of the few labelled samples from the 360 novel classes; and (3) a query set of  synthesised/generated samples for the 360 novel classes (see Sec.~\ref{sect:genfsl}) which augments the support set. Next, our competitive BPL algorithm (to be detailed in Sec.~\ref{sect:adaptfsl}) is run in each learning episode; the outputs (i.e., learned projections) of the episodes are then averaged to obtain the final projection matrix $\mathbf{W}^*$. The novel class labels of test data in the test set are finally predicted with Eq.~(\ref{eq:abplclass}).

From the model overview above, we can see that our SFSP based feature synthesis is now employed as data augmentation to overcome the data sparsity issue in FSL. Importantly, knowledge transfer takes place when the base/source classes are used together with the novel/target class samples (real and synthesised) to learn the projection function. Furthermore, the overall FSL framework is very similar to ZSL in that it is still based on projection learning and once the prototypes are projected, they are used for nearest neighbour search during test. This makes our FSL model drastically different from any existing FSL ones including the related feature synthesised ones \cite{hariharan2017low,Douze_2018_CVPR,Wang_2018_CVPR}. In particular, for the first time, we have presented a unified framework for both zero- and few-shot learning that is applicable to large-scale problems.

\subsection{Feature Synthesis}
\label{sect:genfsl}

\begin{figure}[t]
\vspace{0.1in}
\centering
\includegraphics[width=0.98\columnwidth]{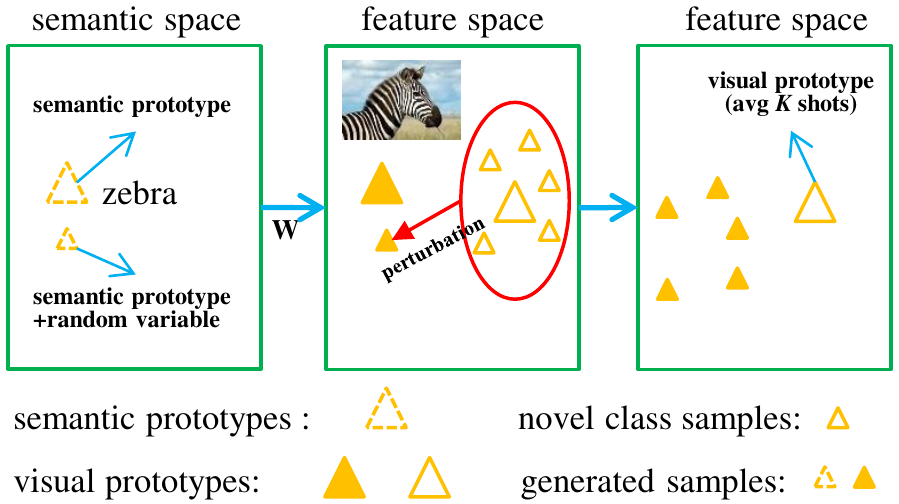}
\vspace{0.02in}
\caption{Illustration of our SFSP strategy for FSL with $K=5$. Only one random variable $\epsilon_i$ is sampled here. }\label{fig:genfsl}
\vspace{-0.0in}
\end{figure}

With a few real samples from each target/novel class, the SFSP strategy originally developed for ZSL needs to be modified. In particular, the SFSP strategy described in Sec.~\ref{sec:ZSL feature synthesis} assumes that the intra-class variations exhibited in seen classes are preserved in a semantically related unseen class. This is a strong assumption that may be invalid in practice. Now with a few real samples from the novel class, the intra-class variation can be more faithfully captured by the real samples. We therefore use the real novel class intra-class variations as offset to perturb the class centre obtained by projecting the class prototype into the visual feature space. Formally, we assume that there are $K$ labelled samples per novel class. Given the initial projection matrix $\mathbf{W}$ learned by our competitive BPL algorithm (setting $\alpha=0$) with all base/seen class samples, novel  class features are computed as follows:
\begin{equation}
\begin{small}
{\mathbf{x}}_i^{(g)} = ({\mathbf{x}}_i^{(u)}- \bar{\mathbf{x}}_j^{(u)}) + \rho \frac{\mathbf{W}(\mathbf{y}_j^{(u)}+\epsilon_i)}{\|\mathbf{W}\|^2_F}, \label{eq:generate1}
\end{small}
\end{equation}
where ${\mathbf{x}}_i^{(u)}$ is a real feature sample from the $j$-th novel class (its semantic prototype is $\mathbf{y}_{j}^{(u)}$), $\bar{\mathbf{x}}_j^{(u)}$ is the feature mean (visual prototype) of the $j$-th  class obtained by averaging the feature vectors of the $K$  samples, $\epsilon_i$ is a random variable in the semantic space to introduce randomness, and $\rho$ is a weight parameter in the range $(0,1)$. The right hand side of Eq.~(\ref{eq:generate1}) has two terms: the first being the intra-class variation and the second the class visual feature mean. This formulation is  thus motivated by the beliefs that (1) The intra-class variation exhibited by the few-shots is the best way to utilise the few-shots, and thus preserved loyally; and (2) the class centres, on the other hand, are based on the semantic prototypes rather than the feature means/prototypes. This is because these prototypes (word vectors in this case) are distilled from human knowledge bases and therefore considered more trustworthy than the few examples.   Such a feature synthesis process is  illustrated in Fig.~\ref{fig:genfsl}. In this work, we empirically synthesise 5 features (i.e. sample $\epsilon_i$ 5 times) for each real sample ${\mathbf{x}}_i^{(u)}$, resulting in $5K$ synthesised samples per novel class.  For each learning episode $h$ ($h=1,...,10$), all generated samples are collected to form a query set $\mathcal{Q}_h = \{\mathbf{x}_{i}^{(g)}: i=1,...,n_h\}$, where $n_h$ denotes the query set size.

\subsection{Model Training}
\label{sect:adaptfsl}

Denote the sample/support/query set of a learning episode  as $\mathcal{D}_s=\{(\mathbf{x}_i^{(s)}, l_i^{(s)}, \mathbf{y}_{l_i^{(s)}}^{(s)}): i=1,...,N_s\}$, $\mathcal{D}_u=\{(\mathbf{x}_i^{(u)}, l_i^{(u)}, \mathbf{y}_{l_i^{(u)}}^{(u)}): i=1,...,N_u\}$,  and $\mathcal{Q}_h = \{\mathbf{x}_i^{(g)}: i=1,...,n_h\}$ respectively, where the notations in $\mathcal{D}_s$ and $\mathcal{D}_u$ are the same as in Sec.~\ref{sect:method}. Our goal is to transfer knowledge from the source $\mathcal{D}_s$ to the target $\mathcal{D}_u$ and $\mathcal{Q}_h$. Deploying Algorithm~\ref{alg:erpl}, now the updates of $\eta_{ij}^{(t)}$ and $\xi_{ij}^{(t)}$ are only related to the small query set, but the updates of $\widehat{\mathbf{A}}^{(t)}$, $\widehat{\mathbf{B}}^{(t)}$, and $\widehat{\mathbf{C}}^{(t)}$ are related to all three sets. By computing $\widehat{\mathbf{A}}_0= \sum_{i=1}^{N_s} \mathbf{x}_i^{(s)} {\mathbf{x}_i^{(s)}}^T $, $\widehat{\mathbf{B}}_0= \sum_{i=1}^{N_s} \mathbf{y}_{l_i^{(s)}}^{(s)} {\mathbf{y}_{l_i^{(s)}}^{(s)}}^T $, and $\widehat{\mathbf{C}}_0= \sum_{i=1}^{N_s} \mathbf{x}_{i}^{(s)} {\mathbf{y}_{l_i^{(s)}}^{(s)}}^T$ in advance before the learning episodes begin, the updates of $\widehat{\mathbf{A}}^{(t)}$, $\widehat{\mathbf{B}}^{(t)}$, and $\widehat{\mathbf{C}}^{(t)}$ become significantly more efficient:
\begin{small}
\begin{eqnarray}
\hspace{-0.25in} && \widehat{\mathbf{A}}^{(t)} = (1\hspace{-0.05in} -\hspace{-0.02in}\alpha_t) \widehat{\mathbf{A}}_0 \hspace{-0.02in}+ \hspace{-0.02in} \alpha_t \widehat{\mathbf{A}}_u \hspace{-0.02in}+\hspace{-0.02in} \alpha_t \hspace{-0.02in} \sum_{i=1}^{n_h} (1\hspace{-0.02in} - \hspace{-0.02in} \mu)\mathbf{x}_i^{(g)} {\mathbf{x}_i^{(g)}}^T\hspace{-0.08in}+ \hspace{-0.02in} \beta I, \label{eq:erplAt0}\\
\hspace{-0.25in} && \widehat{\mathbf{B}}^{(t)} =(1\hspace{-0.05in}- \hspace{-0.02in}\alpha_t)  \widehat{\mathbf{B}}_0 \hspace{-0.02in}+ \hspace{-0.02in} \alpha_t \widehat{\mathbf{B}}_u \hspace{-0.02in}+ \hspace{-0.02in} \alpha_t \hspace{-0.02in} \sum_{i=1}^{n_h} \sum_{j=1}^q \delta_{ij}^{(t)}\mathbf{y}_{j}^{(u)} {\mathbf{y}_{j}^{(u)}}^T \hspace{-0.08in} + \hspace{-0.02in} \beta I, \label{eq:erplBt0} \\
\hspace{-0.25in} && \widehat{\mathbf{C}}^{(t)} = 2(1\hspace{-0.05in} -\hspace{-0.02in} \alpha_t) \widehat{\mathbf{C}}_0  \hspace{-0.02in}+ \hspace{-0.02in} 2\alpha_t \widehat{\mathbf{C}}_u \hspace{-0.02in}+ \hspace{-0.02in} 2\alpha_t \hspace{-0.02in} \sum_{i=1}^{n_h} \sum_{j=1}^q \delta_{ij}^{(t)} \mathbf{x}_{i}^{(g)}{\mathbf{y}_{j}^{(u)}}^T, \label{eq:erplCt0}
\end{eqnarray}
\end{small}
\hspace{-4.6pt}where $\widehat{\mathbf{A}}_u= \sum_{i=1}^{N_u} \mathbf{x}_i^{(u)} {\mathbf{x}_i^{(u)}}^T $, $\widehat{\mathbf{B}}_u= \sum_{i=1}^{N_u} \mathbf{y}_{l_i^{(u)}}^{(u)} {\mathbf{y}_{l_i^{(u)}}^{(u)}}^T $, and $\widehat{\mathbf{C}}_u= \sum_{i=1}^{N_u} \mathbf{x}_{i}^{(u)} {\mathbf{y}_{l_i^{(u)}}^{(u)}}^T$. This is because $\widehat{\mathbf{A}}_u$, $\widehat{\mathbf{B}}_u$, and $\widehat{\mathbf{C}}_u$ are computed using the small support set at a very low cost. With the updated matrices $\widehat{\mathbf{A}}^{(t)}$, $\widehat{\mathbf{B}}^{(t)}$, and $\widehat{\mathbf{C}}^{(t)}$, the best projection matrix can be found efficiently by solving Eq.~(\ref{eq:erpllinear1}). Each learning episode thus has a linear time complexity with respect to the small support/query set size.

\vspace{-0.1cm}
\section{Experiments}

\vspace{-0.0cm}
\subsection{Zero-Shot Learning}
\label{sect:exp:zsr}

\vspace{-0.0cm}
\subsubsection{Datasets and Settings}

\noindent\textbf{Datasets}. Four widely-used benchmark datasets are selected in this paper. Three of them are of medium-size: Animals with Attributes (AwA) \cite{Lampert2014pami}, CUB-200-2011 Birds (CUB) \cite{CUB-200-2011}, and SUN Attribute (SUN) \cite{Patterson2014ijcv}. One large-scale dataset is ILSVRC2012/2010 \cite{Russakovsky2015ImageNet} (ImNet), where the 1,000 classes of ILSVRC2012 are used as seen classes and 360 classes of ILSVRC2010 (not included in ILSVRC2012) are used as unseen classes, as in \cite{Fu2016CVPR}. The details of these datasets are given in Table~\ref{tab:datasets}.

\noindent\textbf{Semantic Spaces}. Two types of semantic spaces are considered for ZSL: attributes are employed to form the semantic space for the three medium-scale datasets, while word vectors are used as semantic representation for the large-scale ImNet dataset. In this paper, we train a skip-gram text model on a corpus of 4.6M Wikipedia documents to obtain the word2vec \cite{mikolov2013distributed} word vectors.

\begin{table}[t]
\vspace{0.2in}
\caption{Details of four benchmark datasets. Notations: `SS' -- semantic space, `SS-D' -- the dimension of semantic space, `A' -- attribute, and `W' -- word vector.}
\label{tab:datasets}
\vspace{-0.1in}
\begin{center}
\begin{small}
\tabcolsep0.3cm
\begin{tabular}{c|c|c|c|c}
\hline
Dataset & \# images & SS & SS-D & \# seen/unseen\\
\hline
AwA & 30,475 & A & 85 & 40/10 \\
CUB & 11,788 & A & 312 & 150/50 \\
SUN & 14,340 & A & 102 & 645/72 \\
\hline
ImNet & 254,000 & W & 1,000 & 1,000/360\\
\hline
\end{tabular}
\end{small}
\end{center}
\vspace{-0.05in}
\end{table}

\noindent\textbf{Visual Features}. All recent ZSL models use the visual features extracted by CNN models \cite{simonyan2014arxiv,szegedy2015cvpr,he2016cvpr}, which are pre-trained on the 1K classes in ILSVRC 2012 \cite{Russakovsky2015ImageNet}. In this paper, for fair comparison, we extract the ResNet101 features \cite{he2016cvpr} for the three medium-scale datasets as in \cite{Xian2017CVPR,Chen2018CVPR} and the GoogLeNet features \cite{szegedy2015cvpr} for the large-scale ImNet dataset as in \cite{Kodirov2017CVPR,Zhang2017cvpr}. Note that the same visual features are used for all compared methods unless stated otherwise.

\noindent\textbf{ZSL Settings}. (1) Pure ZSL: A new `pure' ZSL setting \cite{Xian2017CVPR} is recently proposed to overcome the weakness of the old and standard ZSL setting followed by the majority of prior work. Concretely, most recent ZSL models extract the visual features for the three medium-scale datasets using ImageNet ILSVRC2012 1K classes pretrained CNN models, but the unseen classes in the standard splits overlap with the 1K ImageNet classes. The zero-shot rule is thus violated. Under the pure ZSL setting, the overlapped ImageNet classes are removed from the test set of unseen classes for the new benchmark dataset splits. As for the large-scale ImNet dataset, its ILSVRC2012/2010 split naturally gives a pure ZSL setting. (2) Generalised ZSL: Another ZSL setting that emerges recently \cite{Xian2017CVPR} is the generalised setting under which the test set contains data samples from both seen and unseen classes. This setting is clearly more reflective of real-world applications.

\noindent\textbf{Evaluation Metrics}. (1) Pure ZSL: For the three medium-scale datasets, we compute average per-class top-1 accuracy as in \cite{Xian2017CVPR,Chen2018CVPR}. For the large-scale ImNet dataset, flat hit@5 accuracy is computed over all test samples as in \cite{Frome2013nips,Fu2016CVPR}. (2) Generalised ZSL: Three metrics are defined: 1) $acc_s$ -- average per-class top-1 accuracy of classifying test samples from the seen classes to all classes (both seen and unseen); 2) $acc_u$ -- average per-class top-1 accuracy of classifying test samples from the unseen classes to all classes; 3) HM -- harmonic mean of $acc_s$ and $acc_u$. Many ZSL models have a free parameter that can be tuned to trade off between $acc_s$ and $acc_u$. Therefore, the HM metric gives an overall picture of how a ZSL model evaluated performs under this more challenging yet realstic setting.

\noindent\textbf{Parameter Settings}. Our competitive BPL model  has three free parameters: $\rho\in (0,1)$ in Eq.~(\ref{eq:generate}) for feature synthesis, $\alpha\in (0,1)$ in Step 3 of Algorithm~\ref{alg:erpl}, and $\mu\in (0,1)$ in Step 5 of Algorithm~\ref{alg:erpl}. As in \cite{Kodirov2017CVPR}, the three parameters are selected by class-wise cross-validation on the training set for each benchmark dataset.

\noindent\textbf{Compared Methods}. A wide range of existing ZSL models are selected for comparison. Under each ZSL setting, we focus on the recent and representative ZSL models that have achieved the state-of-the-art results.

\begin{table}[t]
\vspace{0.2in}
\caption{Comparative results  under pure ZSL. %The hit@5 accuracy (\%) is used for ImNet.
 }
\label{tab:pzsl}
\vspace{-0.1in}
\begin{center}
\begin{small}
\tabcolsep0.3cm
\begin{tabular}{l|c|c|c|c
}
\hline
Model &   AwA & CUB &  SUN & ImNet \\
\hline
CMT \cite{socher2013nips}  & 39.5 & 34.6 & 39.9 & -- \\
DeViSE \cite{Frome2013nips}   & 54.2 &  52.0 &  56.5 & 12.8\\
DAP \cite{Lampert2014pami} & 44.1 & 40.0 &  39.9 & -- \\
ConSE \cite{Norouzi14iclr} & 45.6 & 34.3 &  38.8 & 15.5 \\
SSE \cite{Zhang2015iccv}  & 60.1 & 43.9 &  51.5 & -- \\
SJE \cite{Akata2015CVPR}  & 65.6 & 53.9 &  53.7 & -- \\
ALE \cite{akata2016pami} & 59.9 & 54.9 &  58.1 & -- \\
SynC \cite{Changpinyo2016CVPR} & 54.0 & 55.6 &  56.3 & --\\
CVAE \cite{Mishra2017} & 71.4 & 52.1 & 61.7 & 24.7 \\
SAE \cite{Kodirov2017CVPR} &  61.3  &  48.2 & 59.2 & 27.2 \\
DEM \cite{Zhang2017cvpr} & 68.4 & 51.7 & 61.9 &  25.7\\
VZSL \cite{Wang2018AAAI} & -- & -- & -- & 23.1 \\
SP-AEN \cite{Chen2018CVPR} & 58.5 & 55.4 &  59.2 & -- \\
\textcolor{black}{GAN+ALE \cite{Xian2018CVPR}} & 68.2 & 61.5 &  62.1 & -- \\
\hline
\textcolor{black}{Ours}  & \textbf{74.1} & \textbf{61.9}  &  \textbf{63.2}  & \textbf{28.2} \\
\hline
\end{tabular}
\end{small}
\end{center}
\vspace{-0.05in}
\end{table}

\vspace{-0.1cm}
\subsubsection{Comparative Results}

\noindent\textbf{Pure ZSL}. Table~\ref{tab:pzsl} shows that: (1) Our model achieves superior performance on all four datasets, validating the effectiveness of both our SFSP strategy and the robust competitive BPL model. (2) The improvements obtained by our model over the state-of-the-art feature synthesis models \cite{Mishra2017,Xian2018CVPR} are striking. This is despite the fact that our feature synthesis model is much simpler than the deep generative models adopted by \cite{Mishra2017,Xian2018CVPR}. (3) Our model clearly outperforms the recent bidirectional projection learning based alternatives \cite{Kodirov2017CVPR,Wang2018AAAI,Chen2018CVPR}, partly because we additionally exploit feature synthesis and partly due to the fact that competitive learning is introduced in our formulation. (4) On the large-scale ImNet dataset, our model leads to about 2--5\% improvements over the state-of-the-art deep ZSL models \cite{Wang2018AAAI,Mishra2017,Zhang2017cvpr}, demonstrating the scalability of our model to large-scale problems.

\begin{table}[t]
\vspace{0.2in}
\caption{Comparative generalised ZSL results (\%).}
\label{tab:gsrl}
\vspace{-0.1in}
\begin{center}
\begin{small}
\tabcolsep0.02cm
\begin{tabular}{l|ccc|ccc|ccc}
\hline
\multirow{2}{*}{Model}  &    & AwA  &   &  & CUB &  &  &  SUN & \\
\cline{2-10}
&  $acc_s$ & $acc_u$ & HM & $acc_s$ & $acc_u$ & HM & $acc_s$ & $acc_u$ & HM\\
\hline
CMT \cite{socher2013nips} & 86.9 & 8.4 & 15.3 & 60.1 & 4.7 & 8.7 & 28.0 & 8.7 & 13.3\\
DeViSE  \cite{Frome2013nips} & 68.7 & 13.4 & 22.4 & 53.0 & 23.8 & 32.8 & 27.4 & 16.9 & 20.9\\
SSE \cite{Zhang2015iccv}  & 80.5 & 7.0 & 12.9 & 46.9 & 8.5 & 14.4 & 36.4 & 2.1 & 4.0\\
SJE \cite{Akata2015CVPR}  & 74.6 & 11.3 & 19.6 & 59.2 & 23.5 & 33.6 & 30.5 & 14.7 & 19.8\\
LATEM \cite{xian2016latent} &  71.7 & 7.3 & 13.3 & 57.3 & 15.2 & 24.0 &  28.8 &  14.7 & 19.5 \\
ALE \cite{akata2016pami} & 76.1 & 16.8 & 27.5 & 62.8 & 23.7 & 34.4 & 33.1 & 21.8 & 26.3\\
ESZSL \cite{Romera2015icml} &  75.6 & 6.6 & 12.1 & 63.8 & 12.6 & 21.0 & 27.9 & 11.0 & 15.8 \\
SynC \cite{Changpinyo2016CVPR} & 87.3 & 8.9 & 16.2 & \textbf{70.9} & 11.5 & 19.8 & \textbf{43.3} & 7.9 & 13.4\\
SAE \cite{Kodirov2017CVPR} & 71.3 & 31.5 & 43.5 & 36.1 & 28.0 & 31.5 & 25.0 & 15.8 & 19.4\\
DEM \cite{Zhang2017cvpr} & 84.7 & 32.8 & 47.3 &  57.9 & 19.6 & 29.2 & 34.3  & 20.5 & 25.6 \\
SP-AEN \cite{Chen2018CVPR} & \textbf{90.9} & 23.3 & 37.1 & 70.6 & 34.7 & 46.6 & 38.6 & 24.9 & 30.3 \\
\textcolor{black}{GAN+ALE \cite{Xian2018CVPR}} &  57.2 & 47.6 & 52.0 & 59.3 & 40.2 & 47.9  & 31.1 &  41.3 & 35.5 \\
\hline
\textcolor{black}{Ours}  & 66.8 & \textbf{48.8} & \textbf{56.4} & 52.5 & \textbf{47.3} & \textbf{49.8} & 27.9 & \textbf{42.2}  & 33.6  \\
\hline
\end{tabular}
\end{small}
\end{center}
\vspace{-0.05in}
\end{table}

\noindent\textbf{Generalised ZSL}.  It can be observed from Table~\ref{tab:gsrl} that: (1) Different ZSL models have a different trade-off between $acc_u$ and $acc_s$. For example, some models are clearly biased towards the seen class performance (e.g., SynC \cite{Changpinyo2016CVPR} on CUB and  SP-AEN \cite{Chen2018CVPR} on AwA). This is because these models are trained on seen class data only and thus generalise poorly to unseen classes.  Comparing results measured by HM is thus more meaningful. (2) Our model achieves the best overall performance on AwA and CUB, and is very competitive on SUN. This is very impressive, given that our projection learning model takes a simple linear formulation and the feature synthesis is achieved by merely recycling seen class feature vectors by perturbation, avoiding the often painful process of tuning a GAN as in \cite{Xian2018CVPR}. (3) Note that under the generalised setting, our model simply treats the seen and unseen classes equally and performs nearest neighbour search among all projected prototypes given a test data point. Yet, this seems to have achieved a good balance across the seen and unseen class performances, without consistently biasing towards one of them. In contrast, a bias is demonstrated by many compared models, typically towards the seen classes, as expected. This good characteristic is also shared by the feature synthesis based GAN+ALE \cite{Xian2018CVPR}, suggesting that feature synthesis is indeed effective in narrowing the domain gap between the seen and unseen class domains.

\begin{figure}[t]
\vspace{0.05in}
\begin{center}
\includegraphics[width=0.44\textwidth]{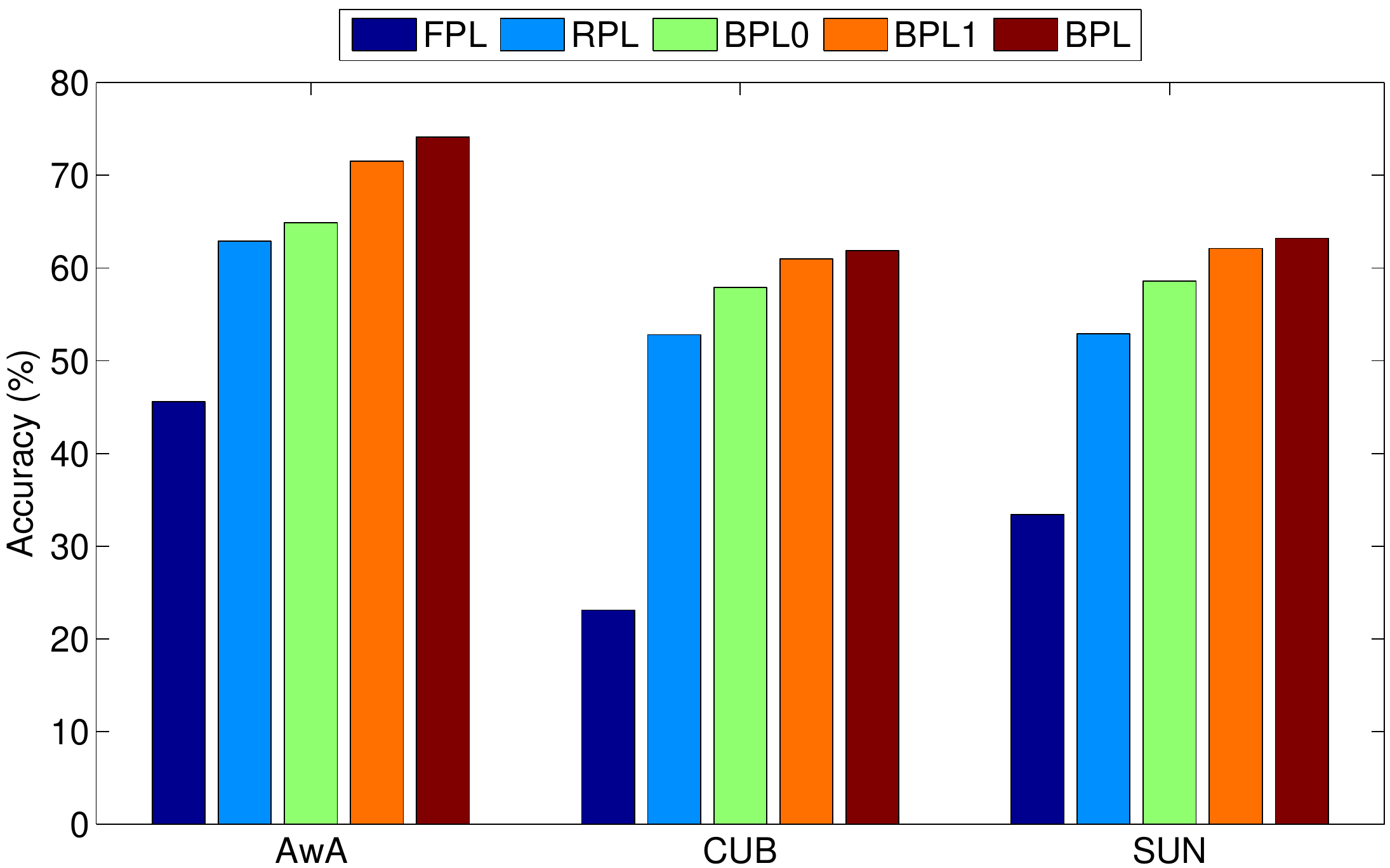}
\end{center}
\vspace{-0.1in}
\caption{Ablation study results on the three medium-scale datasets under pure ZSL. } \label{fig:comp}
\vspace{0.00in}
\end{figure}

\begin{figure*}[t]
\vspace{0.1in}
\begin{small}
\begin{center}
\subfigure[FPL]{
\includegraphics[width=0.18\textwidth,height=0.16\textwidth]{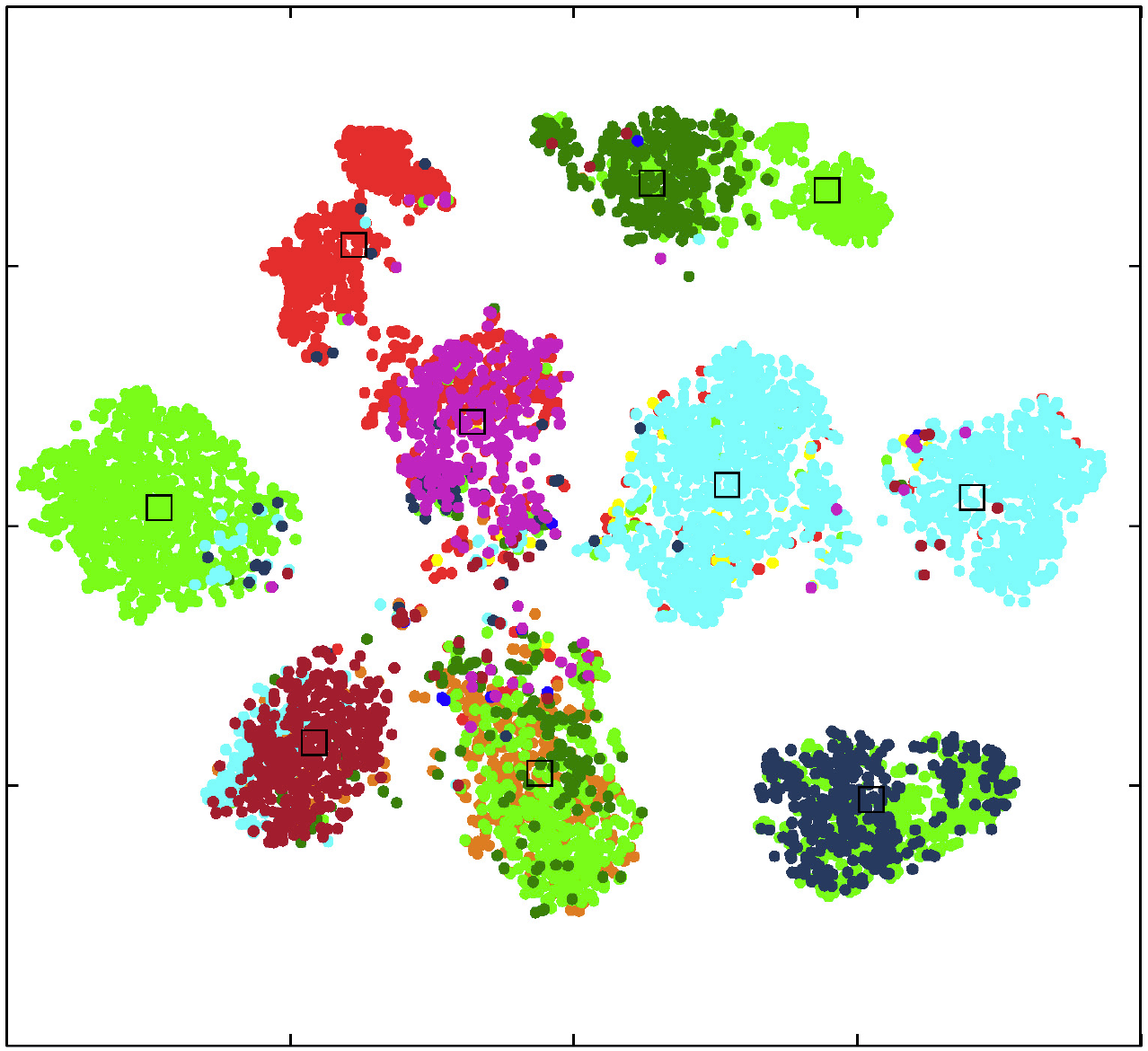}
}%
\subfigure[RPL]{
\includegraphics[width=0.18\textwidth,height=0.16\textwidth]{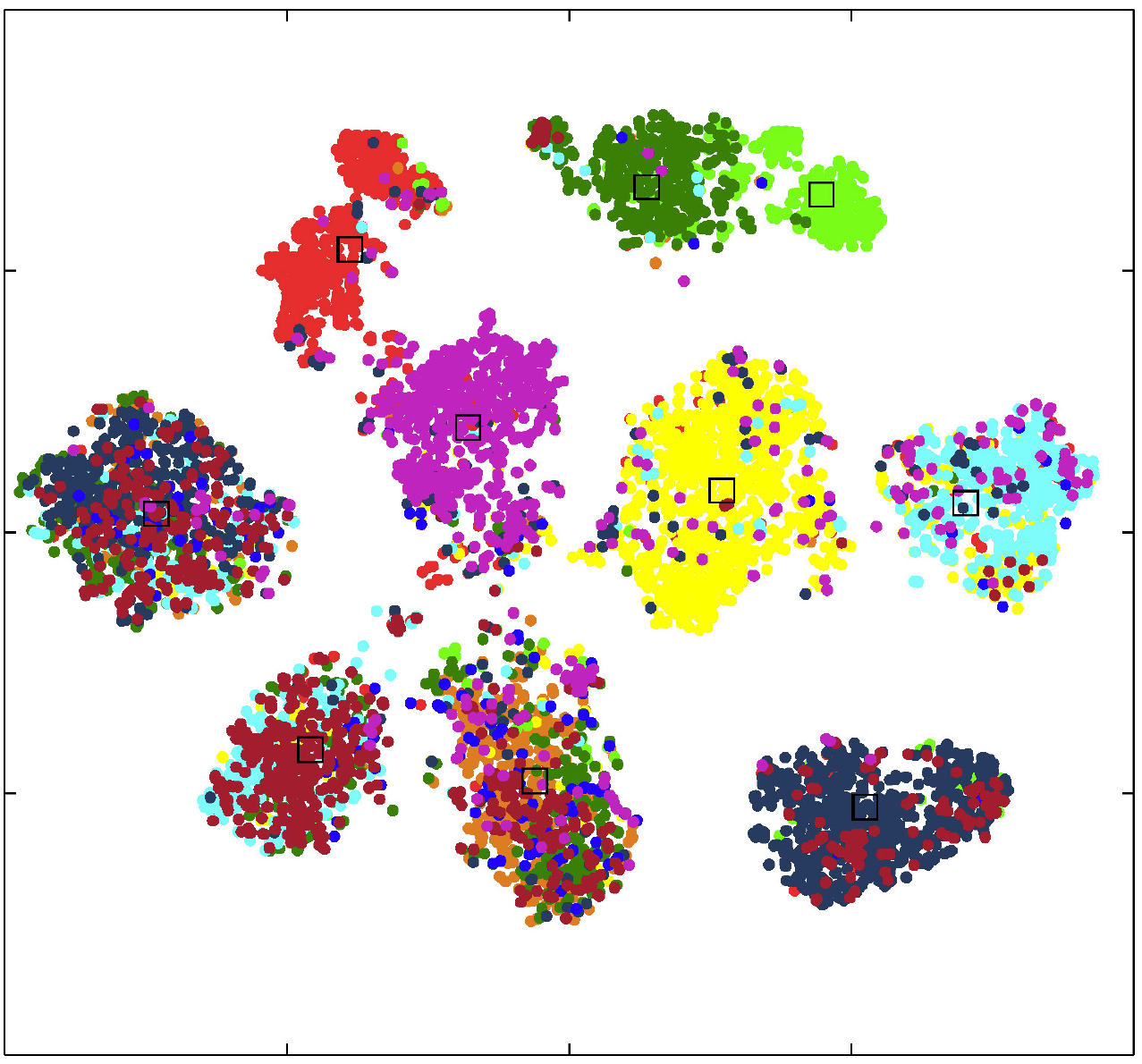}
}%
\subfigure[BPL0]{
\includegraphics[width=0.18\textwidth,height=0.16\textwidth]{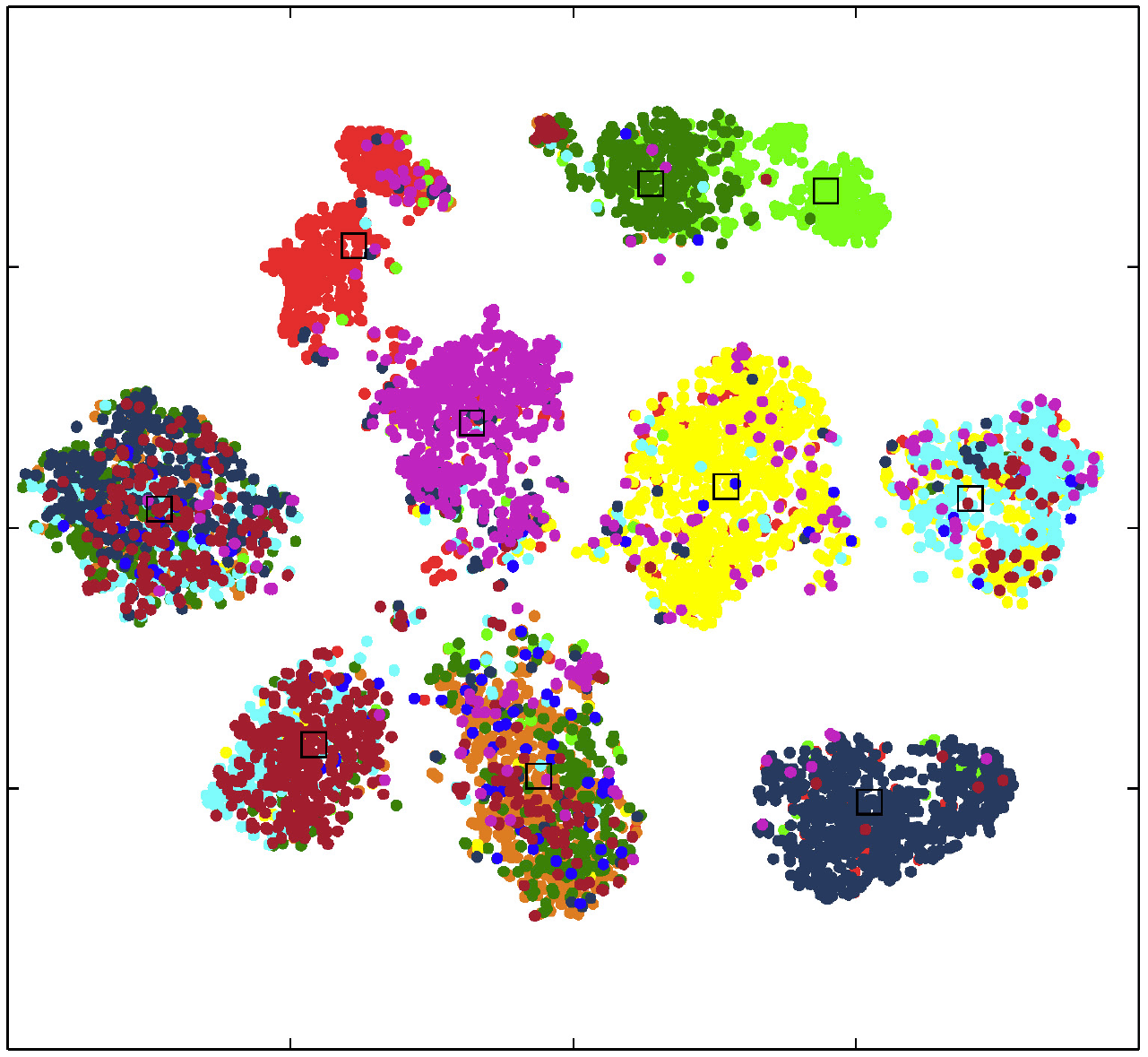}
}%
\subfigure[BPL1]{
\includegraphics[width=0.18\textwidth,height=0.16\textwidth]{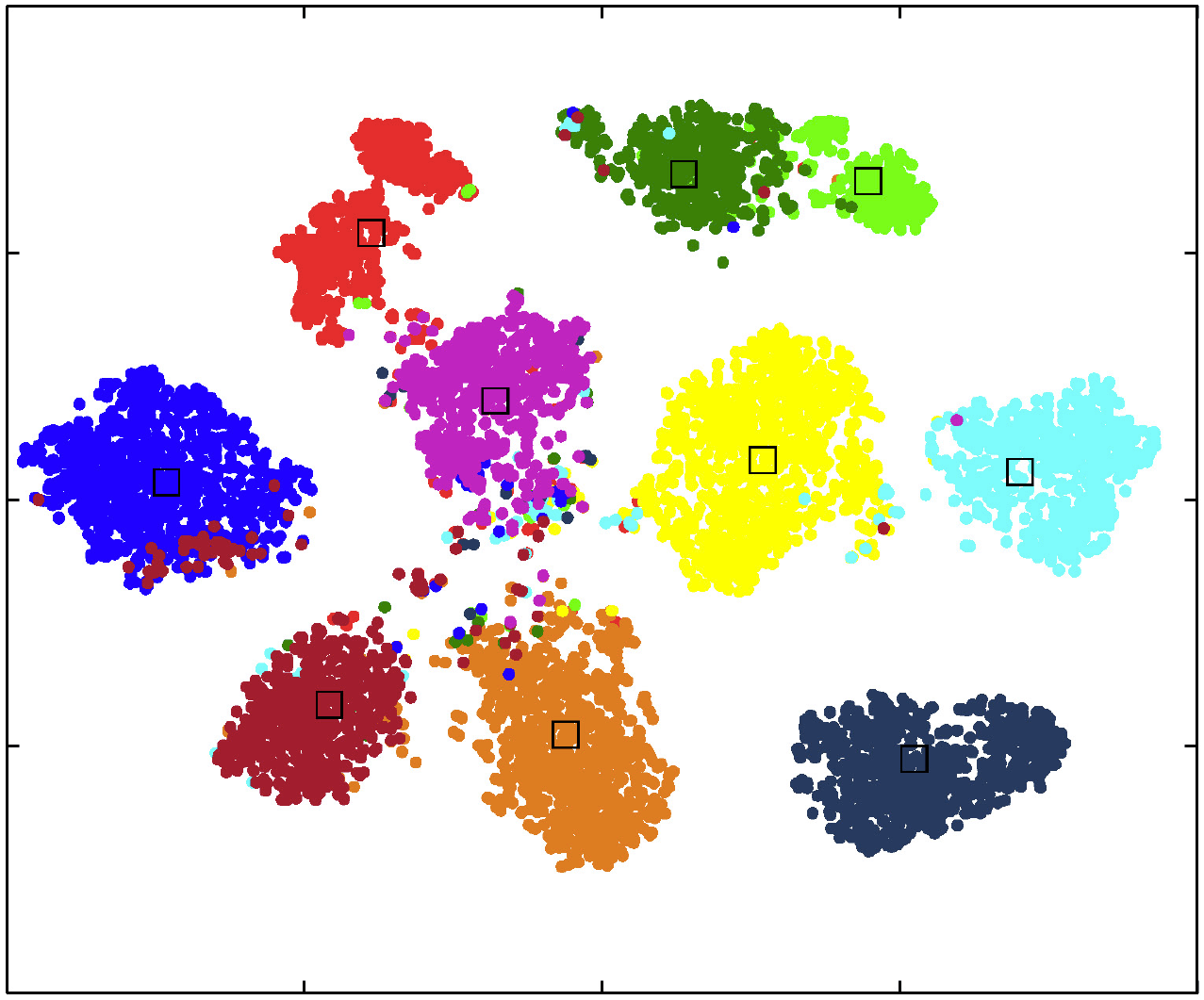}
}%
\subfigure[BPL]{
\includegraphics[width=0.18\textwidth,height=0.16\textwidth]{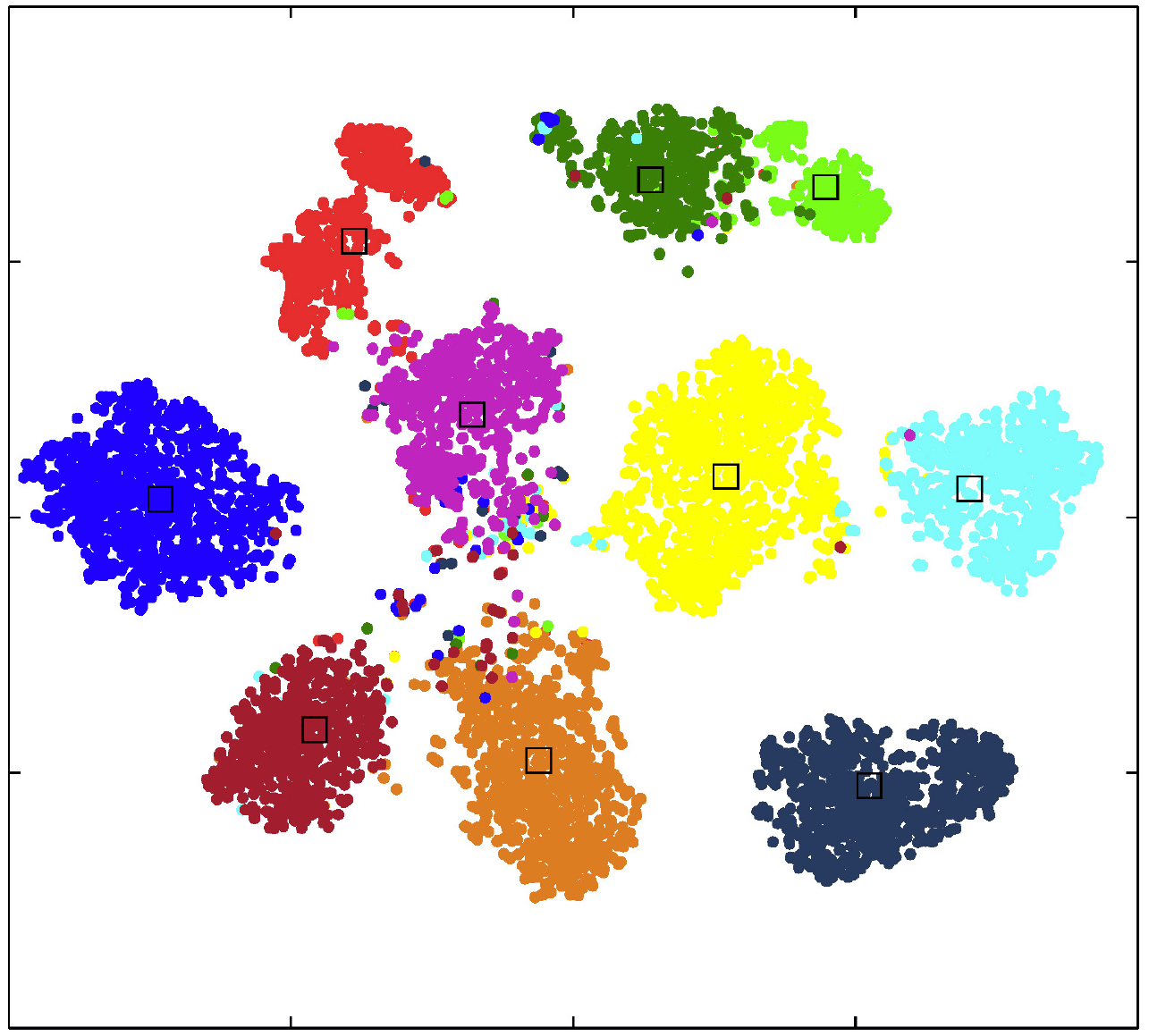}
}
\end{center}
\vspace{-0.12in}
\caption{The tSNE visualisation of the visual features of test unseen class samples from the AwA dataset together with the projected class prototypes. The predicted unseen class labels (marked with different colors) of the test samples are obtained by FPL, RPL, BPL0, BPL1, and BPL under the pure ZSL setting, respectively.}
\label{fig:vis}
\end{small}
\vspace{0.05in}
\end{figure*}

\vspace{-0.1cm}
\subsubsection{Further Evaluations}
\label{sec:further}

\noindent\textbf{Ablation Study}. In this study, various stripped-down versions of our full model (SFSP+competitive BPL, and simply denoted as BPL here) to evaluate the contributions of various key components of the model. Specifically,  (1) When competitive learning is not used for ZSL (i.e. $\mu=0$ in Eq.~(\ref{eq:erplobj}), our model (still with SFSP) is denoted as BPL1. (2) When $\alpha=0$, the BPL1 model further degrades to a conventional BPL model that is trained with seen class samples only, denoted as BPL0. (3) When the forward projection is not considered for ZSL, the BPL0 model becomes the original reverse projection learning model \cite{Shigeto2015}, denoted as RPL; (4) When the projection direction is forward, i.e., from the feature to the semantic space, we have FPL. The results in Fig.~\ref{fig:comp} show that all components contribute to the superior performance of our model: (1) Feature synthesis makes the biggest contribution (see BPL1 vs.~BPL0), resulting in improvements ranging from 3\% to 7\%. (2) \textcolor{black}{ Competitive learning brings in 1--3\% gains (see BPL vs.~BPL1).} (3) Bidirectional projection learning is clearly better than learning along a single direction (see BPL0 vs.~RPL/FPL). Some qualitative results can be seen in Fig.~\ref{fig:vis}. It shows that the test unseen class samples are distributed more compactly and more centred around the unseen class prototypes when more components are included in our model, explaining the  better ZSL performance of the full model.

\noindent\textbf{Alternatives to Competitive Learning}. With our model formulation in Eq.~(\ref{eq:erplobj}), the proposed competitive learning model explicitly deals with the label ambiguity in the synthesised data, and focuses on the best and second-best predicted labels for each synthesised unseen class sample to disambiguate. There are alternatives to this clustering-style unsupervised learning objective. \textcolor{black}{The simplest one is to ignore the label ambiguity, denoted as  `w/o ambiguity handling'. Using this method, each synthesised data point is assigned to the unseen class whose prototype was used as perturbation. That is, $\eta_{ij}^{(t)}=1$ if ${\mathbf{x}}_i^{(g)}$ is labelled as unseen class $j$ and $\eta_{ij}^{(t)}=0$ otherwise (see Eq.~(\ref{eq:subgrad0})). This is essentially the strategy adopted by all existing feature synthesis based ZSL \cite{bucher2017iccv,Mishra2017,Long2017FromZL,Xian2018CVPR,Zhu2018CVPR} or FSL methods \cite{hariharan2017low,Douze_2018_CVPR,Wang_2018_CVPR}, but under the same perturbation framework for fair comparison.   \textcolor{black}{The second alternative performs a different form of competitive learning using  minimum entropy, enforcing that each synthesised data is pulled towards the more likely class and pushed away from the rest}:
\begin{small}
\begin{eqnarray}
&&\hspace{-0.06in}\min_{\mathbf{W}}\hspace{-0.02in}\sum_{i=1}^{N_s} (\|\mathbf{W}^T \mathbf{x}_i^{(s)}\hspace{-0.02in} - \mathbf{y}_{l_i^{(s)}}^{(s)}\|^2_2 + \|\mathbf{x}_i^{(s)}\hspace{-0.02in} - \mathbf{W}\mathbf{y}_{l_i^{(s)}}^{(s)}\|^2_2)  \nonumber\\
&&\hspace{0.15in} + 2\nu \|\mathbf{W}\|^2_F - \gamma\sum_{i=1}^{N_g}\sum_{j=1}^{q} P(j|{\mathbf{x}}_i^{(g)})\mathrm{log}(P(j|{\mathbf{x}}_i^{(g)})),
\label{eq:entropyobj}
\end{eqnarray}
\end{small}
\hspace{-4pt}where $P(j|{\mathbf{x}}_i^{(g)}) =\frac{\exp(-\mathrm{loss}_j({\mathbf{x}}_i^{(g)}))}{\sum_{l} \exp(-\mathrm{loss}_{l}({\mathbf{x}}_i^{(g)}))}$, and $\mathrm{loss}_j({\mathbf{x}}_i^{(g)})= \|\mathbf{W}^T{\mathbf{x}}_i^{(g)} \hspace{-0.02in}- \mathbf{y}_j^{(u)}\|^2_2 + \|{\mathbf{x}}_i^{(g)} \hspace{-0.02in}- \mathbf{W}\mathbf{y}_j^{(u)}\|^2_2$. The above optimisation problem can be solved using gradient descent. The main difference is thus on whether to focus on only the second most likely class or all other classes. Table~\ref{tab:gradopt} shows that our competitive learning model is clearly better than both alternatives. It indicates that (1) dealing with the ambiguity in the synthesised data is critical and (2) focusing on the second best class and the resultant hybrid gradient ascent/descent optimisation lead to more effective projection learning than the descent only formulation.}

\begin{table}[t]
\vspace{0.2in}
\caption{\textcolor{black}{Comparative results (\%) obtained by alternative competitive learning methods under pure ZSL}. }
\label{tab:gradopt}
\vspace{-0.1in}
\begin{center}
\begin{small}
\tabcolsep0.4cm
\begin{tabular}{l|c|c|c}
\hline
Model &   AwA & CUB  & SUN \\
\hline
\textcolor{black}{W/O Amb. Handling}  & 62.9 &  54.9  &  57.8 \\
\textcolor{black}{Minimum Entropy}  &  67.1 &  59.0  &  59.6 \\
\textcolor{black}{Ours}  & \textbf{74.1} & \textbf{61.9}  & \textbf{63.2} \\
\hline
\end{tabular}
\end{small}
\end{center}
\vspace{-0.05in}
\end{table}

\begin{figure*}[t]
\vspace{0.1in}
\begin{small}
\begin{center}
\subfigure[hit@1 accuracy]{
\includegraphics[width=0.42\textwidth]{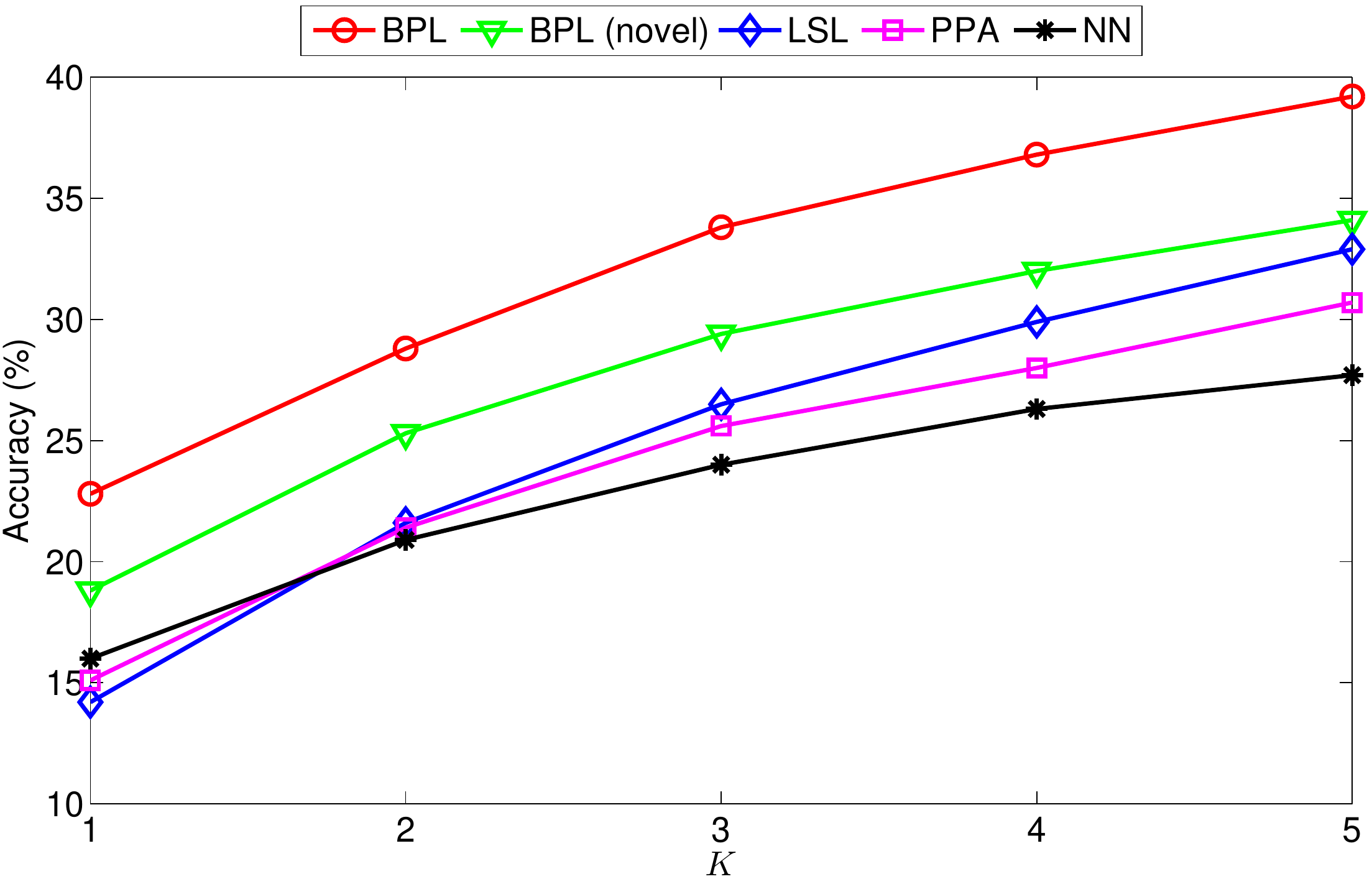}
}%
\hspace{0.05in}
\subfigure[hit@5 accuracy]{
\includegraphics[width=0.42\textwidth]{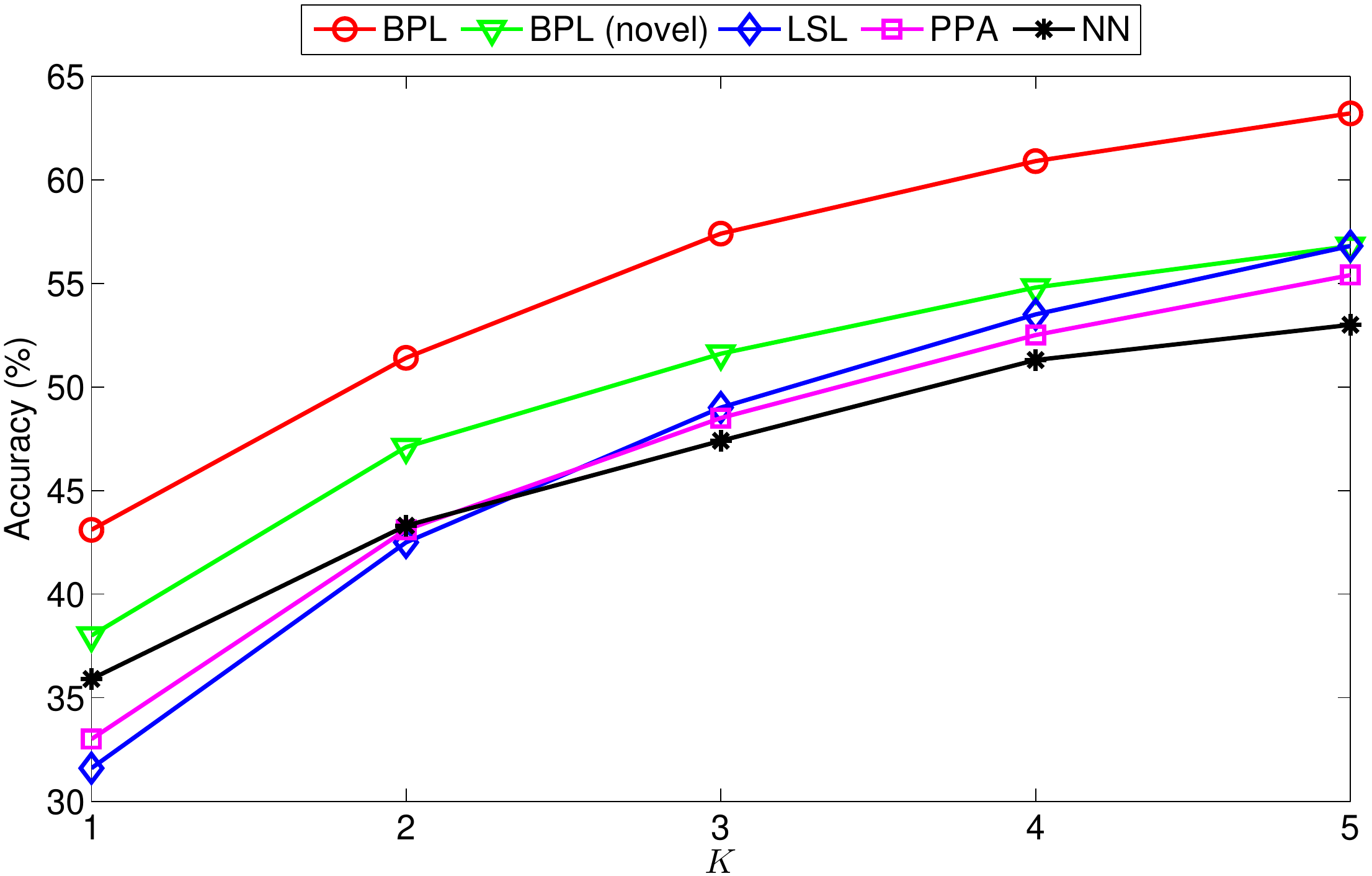}
}
\end{center}
\vspace{-0.15in}
\caption{Few-shot learning results on the large-scale ImNet dataset.}
\label{fig:metafsl1}
\end{small}
\vspace{0.00in}
\end{figure*}

\vspace{-0.1cm}
\subsection{Few-Shot Learning}

\vspace{-0.0cm}
\subsubsection{Large-Scale FSL}
\label{sect:exp:lsfsl}

\noindent\textbf{Dataset and Settings}. We further provide comparative evaluation of our BPL moel under the large-scale FSL setting over the ImNet dataset as described in Sec.~\ref{sect:lsfsl}. The semantic space is the same word vector based one as in ZSL on ImNet, while the visual features are extracted using the ImageNet ILSVRC2012 1K classes pretrained ResNet50 \cite{he2016cvpr}. We compare our full model (BPL) with four models: (1) NN --  nearest neighbour (NN) search baseline performed in the feature space using $K$ samples per novel class as the references. The knowledge transfer is via the feature space learned using the seen/base classes only. (2) BPL (novel) -- our BPL model trained only with few novel class samples, where no knowledge transfer from the base classes is explicitly modelled. (3) LSL -- the low-shot learning (LSL) model \cite{hariharan2017low}; (4) PPA -- the parameter prediction with activations (PPA) model \cite{qiao2017few}.  Note that the pretrained ResNet50 on the ImageNet ILSVRC2012 1K classes (i.e., the base classes) is employed for network initialisation or feature exaction for all compared models. The hit@1 and hit@5 accuracies are computed over all test samples from the 360 novel classes and used as the evaluation metrics.

\noindent\textbf{Comparative Results}. The comparative results of large-scale FSL are presented in Fig.~\ref{fig:metafsl1}. It can be seen that: (1) Our full model (BPL) clearly outperforms the state-of-the-art large-scale FSL methods -- LSL and PPA, and the improvements are more significant with smaller $K$ value. (2) As shown in Table \ref{tab:pzsl}, with zero-shot and class prototypes only, our model obtains a hit@5 accuracy of 28.2\%. Fig.~\ref{fig:metafsl1}(b) shows that with only 1-shot, we obtain a massive boost of 15\%. This clearly demonstrates the effectiveness of our model in utilising few training samples from novel classes thanks to feature synthesis and robust projection learning.  (3) By employing the proposed SFSP and competitive BPL on the novel classes only, our model (BPL (novel)) is still clearly better than LSL and PPA. In the case, the transfer learning is via only the feature space learned using the base classes. Comparing our full model with BPL (novel), the gap suggests that further knoweldge transfer by joint projection learning  boosts the performance by a big margin. (4) The baseline NN method is surprisingly competitive, even beating the state-of-the-art when $K=1$. Note that both LSL and PPA also freeze the base class pretrained ResNet50 for visual feature extraction. This result thus suggests that most of the knowledge transfer is done by this initial step and the  actual transfer learning  methods proposed in LSL and PPA are not effective with low $K$ values.  (5)  Note that none of the five compared methods are meta-learning based. We also evaluated  a number of standard meta-learning methods that provide public source codes, such as prototypical net \cite{snell2017prototypical}. They all fail miserably on this large-scale dataset, indicating their clear shortcomings in scalability.

\noindent\textbf{Alternative Feature Synthesis Strategy}. As discussed in Sec.~\ref{sect:genfsl}, the proposed SFSP strategies for ZSL and FSL, though similar in principle, differs in whether the base class features are used for novel class feature synthesis (comparing Eq.~(\ref{eq:generate1}) with Eq.~(\ref{eq:generate})). There is another subtle difference that makes our FSL model more scalable. If we were to use the base class samples for synthesis, in real-world applications, storing these feature vectors in memory could create a big bottleneck. In contrast, with the strategy in Eq.~(\ref{eq:generate1}), we only need to load the three matrices computed using the base class samples, namely $\widehat{\mathbf{A}}_0$, $\widehat{\mathbf{B}}_0$, and $\widehat{\mathbf{C}}_0$, with a much smaller memory footprint.  Table~\ref{tab:datagen} compares the two strategies (SFSP1 in Eq.~(\ref{eq:generate1})  and SFSP2 in Eq.~(\ref{eq:generate})). It can be seen that SFSP1 is not only more memory efficient, it is also slightly better in performance, suggesting that the intra-class variations from the real few-shot are more trustful reflections of the real variations than the ones transferred from base classes.

\begin{table}[t]
\vspace{0.2in}
\caption{Comparative results (\%) of different feature synthesis strategies for FSL on ImNet.}
\label{tab:datagen}
\vspace{-0.1in}
\begin{center}
\begin{small}
\tabcolsep0.45cm
\begin{tabular}{l|cc|cc}
\hline
\multirow{2}{*}{$K$}  &  \multicolumn{2}{c|}{hit@1 accuracy}  &   \multicolumn{2}{c}{hit@5 accuracy} \\
\cline{2-5}
&  SFSP1 & SFSP2 & SFSP1 & SFSP2 \\
\hline
1 & {\bf 22.8} & 22.7 & {\bf 43.1} & 43.0 \\
2 & {\bf 28.8} & {\bf 28.8} & {\bf 51.4} & 51.3 \\
3 & {\bf 33.8} & 33.7 & {\bf 57.4} & 57.3 \\
4 & {\bf 36.8} & 36.7 & {\bf 60.9} & 60.8 \\
5 & {\bf 39.2} & {\bf 39.2} & {\bf 63.2} & 63.0 \\
\hline
\end{tabular}
\end{small}
\end{center}
\vspace{-0.1in}
\end{table}

\begin{table}[t]
\vspace{0.2in}
\caption{Runtime (mins) comparison for FSL on ImNet.}
\label{tab:fsltime}
\vspace{-0.1in}
\begin{center}
\begin{small}
\tabcolsep0.21cm
\begin{tabular}{l|c|c|c|c}
\hline
Method &  Ours (SFSP1) &  Ours (SFSP2) & LSL & PPA \\
\hline
Runtime  &   {\bf 1.2}  &   2.3  &   23.0  &  1,300.0 \\
\hline
\end{tabular}
\end{small}
\end{center}
\vspace{-0.2in}
\end{table}

\noindent\textbf{Runtime Comparison}. \textcolor{black}{The training times (minutes) by different large-scale FSL methods are given in Table~\ref{tab:fsltime}, which are obtained from the same PC platform with two 2.40 GHz CPUs, 96 GB RAM, and one Tesla K80 GPU. Our model with SFSP1 is shown to be the most efficient for large-scale FSL, and is orders of magnitude more efficient than LSL and PPA.}

\subsubsection{Small-Scale FSL}

\noindent\textbf{Dataset and Settings}. Most published FSL models are based on meta-learning  \cite{finn2017model,snell2017prototypical,sung2017learning,qiao2017few} and only report results on small-scale problems. To compare our model with more published work,  we use a popular benchmark small-scale FSL dataset, the mini-ImageNet as in \cite{snell2017prototypical}.  This dataset consists of 100 ImageNet classes (80 for base and 20 novel), which is 10-times smaller than the large-scale ImNet. The same semantic space is used, while the visual features are extracted with two CNN models trained from scratch with the training set of mini-ImageNet: (1) Simple -- four conventional blocks as in \cite{snell2017prototypical}; (2) WRN -- wide residual networks \cite{Zagoruyko2016WRN} as in \cite{qiao2017few}. The standard FSL setting is followed: The 5-way accuracy is computed by randomly selecting 5 classes from the 20 novel classes for each test trial, and the average 5-way accuracy over 600 test trials is then used as the evaluation metric.

\noindent\textbf{Comparative Results}. From Table~\ref{tab:metafsl}, we can make the following observations:  (1) Our BPL method achieves state-of-the-art performance under the 5-way 5-shot setting and competitive results under 5-way 1-shot. This suggests that our BPL method is suitable not only for the large-scale FSL but also for the small-scale one. (2) Both ours and PPA are much better, compared with the rest meta-learning based methods. (3) This time, NN is weak. This is expected as the feature extraction CNN trained on the base classes is weak due to the small training set size, leaving scopes for the meta-learning based approaches to take effect. But this knowledge transfer route via feature extraction CNN becomes much stronger when more base classes are made available, as shown in Fig.~\ref{fig:metafsl1}, making NN hard to beat for large-scale FSL. (4) Stronger visual features extracted for FSL yield significantly better results (WRN vs. Simple for PPA and ours).

\begin{table}[t]
\vspace{0.2in}
\caption{Comparative results (\%) with 95\% confidence intervals for small-scale FSL on mini-ImageNet. }
\label{tab:metafsl}
\vspace{-0.1in}
\begin{center}
\begin{small}
\tabcolsep0.1cm
\begin{tabular}{l|l|c|c}
\hline
Model & CNN & 1 shot & 5 shot \\
\hline
Nearest Neighbor & Simple &  41.08$\pm$0.70 & 51.04$\pm$0.65 \\
Matching Net \cite{vinyals2016matching} & Simple &  43.56$\pm$0.84 & 55.31$\pm$0.73\\
Meta-Learn LSTM \cite{ravi2017optimization} & Simple &  43.44$\pm$0.77 & 60.60$\pm$0.71\\
MAML \cite{finn2017model}  & Simple &  48.70$\pm$1.84  & 63.11$\pm$0.92   \\
Prototypical Net \cite{snell2017prototypical}  & Simple & 49.42$\pm$0.78 & 68.20$\pm$0.66 \\
mAP-SSVM  \cite{triantafillou2017nips} & Simple & 50.32$\pm$0.80 & 63.94$\pm$0.72 \\
Relation Net \cite{sung2017learning} & Simple & 50.44$\pm$0.82 & 65.32$\pm$0.70 \\
SNAIL \cite{mishra2018simple} & ResNet20 & 55.71$\pm$0.99 &  68.88$\pm$0.92 \\
PPA \cite{qiao2017few} & Simple &  54.53$\pm$0.40 & 67.87$\pm$0.20 \\
PPA \cite{qiao2017few} & WRN & \textbf{59.60$\pm$0.41} & 73.74$\pm$0.19 \\
\hline
Ours & Simple & 54.20$\pm$0.58 & 65.28$\pm$0.59 \\
Ours & WRN & 58.53$\pm$0.82 & \textbf{75.62$\pm$0.61} \\
\hline
\end{tabular}
\end{small}
\end{center}
\vspace{-0.1in}
\end{table}

\vspace{-0.1cm}
\section{Conclusion}

We have proposed a novel ZSL model based on semantic feature synthesis by perturbation and robust  bidirectional project learning with a generalised competitive learning formulation. An efficient iterative algorithm has been developed for model optimisation, followed by rigorous theoretic algorithm analysis. Importantly, the model can be easily extended to large-scale FSL, making our model a unified solution to large-scale visual recognition with insufficient training samples.   Extensive experiments have been carried out to provide strong evidence that the proposed model is more effective than the state-of-the-art alternatives, especially under the large-scale settings. A number of directions are worth further investigation. First, in the current framework, our focus is on developing a robust projection learning model, with the visual features extracted with a pre-trained and fixed CNN model. It is possible to integrate the two models into a single one for joint optimisation.  More efforts are also needed to investigate how to formulate both competitive learning and bidirectional projection learning in end-to-end training framework. Second, in a more practical scenario, each seen class  may also have only a few labelled samples. This provides additional challenges that cannot be addressed using the current model.  Finally, it is worth pointing out that the gradient-based optimisation algorithm introduced in Sec.~\ref{sect:method} is by no means restricted to the ZSL problem -- many other vision problems need to deal with a mix of min-min/max-min problems, for which our hybrid gradient descent/ascent formulation can be applied. Part of the current efforts thus also include the generalisation of the proposed model to solve other vision problems (e.g., social image classification and cross-modal image retrieval).

\vspace{-0.1cm}
\section*{Acknowledgements}

This work was partially supported by National Natural Science Foundation of China (61573363), 973 Program of China (2015CB352502), the Fundamental Research Funds for the Central Universities and the Research Funds of Renmin University of China (15XNLQ01), and European Research Council FP7 Project SUNNY (313243).


\end{document}